\documentclass[twoside]{article}

\usepackage[accepted]{aistats2024}
\usepackage[hidelinks]{hyperref}
\usepackage{self_defined_func}

\begin{document}

%

%

\twocolumn[

\aistatstitle{TransFusion: Covariate-Shift Robust Transfer Learning for High-Dimensional Regression}

\aistatsauthor{ Zelin He \And Ying Sun \And  Jingyuan Liu \And Runze Li }

\aistatsaddress{ Dept. of Statistics \\ Penn State Univ.
 \And  School of EECS, \\ Penn State Univ. \And Dept. of Statistics \\ and Data Science, \\  SOE, Xiamen Univ.  \And Dept. of Statistics, \\ 
 Penn State Univ.} ]

\begin{abstract}
 The main challenge that sets transfer learning apart from traditional supervised learning is the distribution shift, reflected as the shift between the source and target models and that between the marginal covariate distributions.
In this work, we tackle model shifts in the presence of covariate shifts in the high-dimensional regression setting. Specifically, we propose a two-step method with a novel fused-regularizer that effectively leverages samples from source tasks to improve the learning performance on a target task with limited samples. Nonasymptotic bound is provided for the estimation error of the target model, showing the robustness of the proposed method to covariate shifts. We further establish conditions under which the estimator is minimax-optimal. Additionally, we extend the method to a distributed setting, allowing for a pretraining-finetuning strategy, requiring just one round of communication while retaining the estimation rate of the centralized version. Numerical tests validate our theory, highlighting the method's robustness to covariate shifts.
\end{abstract}

Transfer learning is a technique that leverages knowledge from source tasks to improve learning performance in a related but possibly different target task
 \citep{torrey2010transfer}.
 In this paper, we consider the high-dimensional setting where the target sample size is much smaller than the number of features. In this context, applying transfer learning techniques to extract information from a larger pool of source samples can be particularly beneficial in identifying the model parameters.
For example, in genetic studies of rare diseases,  transferring information from larger, related set of studies could uncover highly disease-relevant genetic patterns \citep{taroni2019multiplier}.

In contrast to learning from i.i.d. samples, a fundamental challenge in transfer learning is handling the distribution shifts between the source sample $(\bX_{S}, \by_{S})$ and the target sample $(\bX_{T}, \by_{T})$ \citep{pan2009survey}. The discrepancy between the distributions typically shows in two ways: 1) \textit{model shift: $P(\by_{S}|\bX) \neq P(\by_{T}|\bX)$}, indicating a shift in the learning models, and 2) \textit{covariate shift: $P(\bX_{S}) \neq P(\bX_{T})$}, indicating a shift in the marginal covariate distributions. In either case, models achieving small training errors on the source tasks may experience high risks on the target task \citep{lu2020harder}. Therefore, to improve the learning performance of the target model $P(\by_{T}|\bX)$ using knowledge from the source samples, one should not only capture and correct the model shift but also be robust to the covariate shifts. {In the high-dimensional setting, handling such differences becomes even more difficult due to accumulated noise and limited samples \citep{fan2020statistical}}. This leads to the following question:

How to tackle \textit{model shifts} in \textit{high-dimensional} transfer learning  while being robust to \textit{covariate shifts}?

\addtocounter{footnote}{1} 
\footnotetext{Correspondence to: J Liu <jingyuan@xmu.edu.cn>.}
Apart from the challenge brought by the distribution shift, modern learning problems often involve datasets distributed across multiple computing nodes. In such a scenario, performing centralized training by pooling all the raw data in a single machine can be undesirable due to storage, communication, and privacy issues. This situation prompts another question:

How to transfer knowledge from \textit{distributed} source datasets in a \textit{communication-efficient} manner?

This paper proposes a solution for the above two questions for high-dimensional linear regression with $K$ source tasks, where the target model is $p$-dimensional with sparsity level $s$. 
Our contributions are:\\[1ex]
\noindent$\bullet$ \textbf{Covariate-Shift Robust Regularizer.} We propose a novel fused-regularizer achieving two purposes: it promotes sparse solutions for the high-dimensional model parameter while simultaneously capturing model shifts between source and target datasets. Our theoretical results further show that this regularizer can separate model shifts from shared patterns in a robust manner under covariate shifts.\\[1ex]
\noindent$\bullet$ \textbf{Optimal Estimation Procedure.} Leveraging the proposed regularizer, we introduce a two-step procedure termed TransFusion. When the source tasks are sufficiently diverse, we show applying the first step on the source and target tasks jointly suffices to yield a fast rate of $ O(\frac{s \log p}{n_{T} + Kn_{S}} + \bar{h}\sqrt{\frac{\log p}{n_{S}}})$, where $n_T$ is the target sample size, $n_S$ is the source sample size, and $\bar{h}$ measures  task similarity. The rate significantly improves over the one achieved on target task without transfer learning, i.e., the rate of $O(\frac{s \log p}{n_{T}})$, {when $n_{S}\gg \bar{h}n_{T}^2$}. For cases that do not meet the diversity criteria, TransFusion incorporates a second step refining the estimate on the target task, ensuring a rate of {$O(\frac{s \log p}{n_{T} + Kn_{S}} + \bar{h}\sqrt{\frac{\log p}{n_{T}}} \wedge \bar{h}^2)$}, which is  minimax-optimal when {$p \gg s$} and $\bar{h}$ is relatively small. \\[1ex]
\noindent$\bullet$ \textbf{Efficient Distributed Learning.} We develop a distributed variant of our method, termed D-TransFusion, requiring only one-shot communication of the pre-trained local models from source tasks nodes to target task node, significantly reducing communication overhead. More importantly, it offers the flexibility to quickly adapt the models to different downstream tasks while avoiding training from scratch. We further show that when the source sample size $n_{S}$ is sufficiently large, D-TransFusion achieves the optimal statistical rate, matching its centralized counterpart.\vspace{0.2cm}


\noindent{\bf Related Works:}
This paper develops transfer learning methods for high-dimensional regression problems under both model and covariate shifts. Related works can be broadly divided into the following categories.\\[1ex]
\noindent\textbf{Domain Adaptation} methods primarily focus on handling covariate shifts, usually assuming the underlying models remain the same (\cite{quinonero2008dataset}, \cite{redko2020survey}). 
{One prevalent approach in this category focuses on aligning the source and target covariate distributions by learning domain-invariant representations (\cite{redko2020survey}, \cite{mansour2009domain}, \cite{cortes2011domain}, \cite{cortes2014domain}). Another line of research involves correcting estimators to address covariate shifts, often using the importance weighting (\cite{quinonero2008dataset}, \cite{sugiyama2012machine}, \cite{chen2016robust}). 
In contrast, we explicitly address model shifts and aim for robustness to covariate shifts.}

\textbf{Multitask learning}
 aims to handle model shifts across multiple tasks and learn shared features to improve the performance of each task \citep{pan2009survey}. In regression settings, regularization techniques are often employed to promote information transfer. Examples include the Frobenius and spectral norm (\cite{argyriou2007spectral}, \cite{tian2023repre}), mixed $\ell_{2,1}$ norm \citep{lounici2009taking}, hard-thresholding \citep{huang2023optimal}, 
and the total variation norm (\cite{li2019spatial}, \cite{zhang2022learning}, \cite{tang2016fused}). These works typically require all tasks to have a comparable sample size and emphasize overall task performance. Therefore, they are not directly applicable to transfer learning problems where the target task, often with far fewer samples, is the primary focus.

\noindent\textbf{Transfer Learning} 
has been intensively studied under regression settings (\cite{du2017hypothesis}, \cite{lei2021near}, Lin and Reimherr (\citeyear{lin2022transfer};\citeyear{lin2024smoothness})). However, most works are restricted to low-dimensional problems. {Recently, transfer learning in the high-dimensional regression settings has been studied in \cite{takada2020transfer}, \cite{bastani2021predicting}, \cite{li2022transfer} and \cite{tian2022transfer}. These works, however, deal with scenarios with only a single source or are sensitive to covariate shifts across multiple sources, and their learning accuracy degrades quickly if such shifts are severe.} More recent works such as \cite{li2023estimation} and \cite{liu2023unified} attempt to mitigate the impact of covariate shifts. However, these methods either rely on strong assumptions or are computationally demanding. Specifically, \cite{li2023estimation} established the convergence rate of the proposed estimator assuming the empirical loss function in the high-dimensional setting is smooth, and computing the estimator requires solving a nonsmooth optimization problem with multiple constraints, {while \cite{liu2023unified} assumes the target sample has a comparable size as the source sample}. In contrast, theoretical guarantees of our method are established under weaker, more practical conditions, and numerically it can be computed efficiently using algorithms such as iterative soft thresholding.

\noindent{\bf Notation:}
We use bold upper- and lowercase letters for matrices and  vectors, respectively. For a matrix $\mathbf{A} \in \mathbb{R}^{m \times n}$, we denote its $(i, j)$-th element  by $\mathbf{A}_{ij}$,   maximum eigenvalue by $\Lambda_{\max}(\mathbf{A})$, and  minimum eigenvalue by $\Lambda_{\min}(\mathbf{A})$.  We let $a \vee b$ denote $\max \{a, b\}$ and $a \wedge b$ denote $\min \{a, b\}$. We use $c, c_0, c_1, \ldots$ to denote generic constants independent of $n$, $p$ and $K$.
Let $a_n=O\left(b_n\right)$ and $a_n \lesssim b_n$ denote $\left|a_n / b_n\right| \leq c$ for some constant $c$ when $n$ is large enough; $a_n = o\left(b_n\right)$ and $b_n \gg a_n$ if $a_{n} = O(c_{n}b_{n})$ for some $c_{n} \to 0$; $a \asymp b$ if $a = O(b)$ and $b = O(a)$.


\section{Preliminaries}
\label{sec:preliminaries}
We consider a transfer learning problem involving one target task and $K$ source tasks. For the target task, we observe a sample $(\bXz, \byz)$ generated from the target model
$$
\by^{(0)}_{i}=\left(\bXz_{i \cdot}\right)^{\top} \bz +\bez_{i}, \quad i=1, \ldots, n_{T},
$$
where $\bz \in \mathbb{R}^{p}$ is the parameter of primary interest and $\bez_{i}$ is the observation noise. We focus on a high-dimensional scenario where the dimension $p$ is much larger than the target sample size $n_{T}$, yet the ground truth $\bz$ is a sparse vector with $s:=\|\bz\|_{0}$ nonzero elements, which is much smaller than $p$, {i.e., $p \gg s$.}

In addition to the target sample, we also have access to $K$ source samples $\{(\bXk, \byk)\}_{k 
=1}^K$, generated from the source model
$$
\by^{(k)}_{i}=\left(\bXk_{i \cdot}\right)^{\top} \bk +\bek_{i}, \ i=1, \ldots, n_{S}, \ k = 1, \ldots, K.
$$
For the $k$-th source model, $\bk \in \mathbb{R}^{p}$ is the unknown  source task-specific parameter, and $\bek_{i}$ accounts for the observation noise. For simplicity, we assume the source samples have the same size $n_S$. 

Our goal is to estimate $\bz$ using both the target and source samples under the challenging scenario where distributions of the samples are heterogeneous, as characterized by both model and covariate shift described next.\\[1ex]
\textbf{Model Shift.} In our context, the model shift is the situation where each source model differs from the target model, and is measured by $\bdk:= \bk - \bz$ for $1\le k\le K$. Throughout the paper, we refer $\bdk$ as the ``parameter contrast'' or ``task-specific signal''. A source task is considered informative for transfer learning if $\bdk$ is relatively small.  Formally, 
let $\boldsymbol{\beta}:=((\bz)^{\top}, (\boldsymbol{\beta}^{(1)})^{\top}, \dots, (\boldsymbol{\beta}^{(K)})^{\top})^{\top} \in \mathbb{R}^{(K+1)p}$, we assume $\bb$ belongs to the following parameter space
\begin{align}\label{def:parameter_space}
\Theta(s, \bh):=\left\{\boldsymbol{\beta}:  \|\bz\|_0 \leq s, \|\bk - \bz\|_1 \leq h_{k}\right\},
\end{align}
with $\mathbf{h}:=(h_{1}, \dots, h_{K})^{\top}$. 
In~\eqref{def:parameter_space}, the informative level of the $k$-th source task is quantified by the $\ell_{1}$-sparsity of $\bdk$, and is upper bounded by a factor $h_k \geq 0$. 
\begin{remark}
    {We choose an $\ell_1$-sparse constraint for the high-dimensional contrast $\bk - \bz$, as it aligns well with practical applications where model shifts typically spread over multiple dimensions but their overall magnitude does not grow too fast.} The results in the paper can be naturally extended to a general $\ell_{q}$-sparse case for $q \in [0,1]$.
\end{remark}

\textbf{Covariate Shift.} In addition to the shift in the model parameters, we also consider the covariate shift, defined as the difference in the distributions of $\bXk_{i \cdot}$s across the tasks. In this work, we only impose the following mild tail condition on the distribution of $\bXk_{i \cdot}$s but allow other distribution characteristics, such as the covariance structures, to vary across different tasks.
\begin{assumption}[Sub-Gaussian designs]
\label{A1}
For any $0 \leq k \leq K$, $\bXk_{i \cdot}$s are independent sub-Gaussian random vectors with mean zero and covariance $\Sigk$. Furthermore, 
    there exists some universal constant $c$ such that $1/c \leq \min_{0 \leq k \leq K}\Lambda_{\min}(\Sigk) \leq \max_{0 \leq k \leq K}\Lambda_{\max}(\Sigk) \leq c$. 
\end{assumption}

Finally, we assume that the random noises follow independent Gaussian distributions, a typical assumption for high-dimensional regression analysis.
\begin{assumption}[Gaussian random errors]
\label{A2}
    For all $0 \leq k \leq K$, the $\bek_i$s are independent Gaussian random variables with zero mean and uniformly upper bounded variance, and are independent of $\bXk$s.
\end{assumption}

\section{Covariate-Shift Robust Transfer Learning}
\label{sec:TransFusion}
We now introduce a method called \textbf{TransFusion} (\textbf{Trans}fer Learning with a \textbf{Fu}sed-Regulariza\textbf{tion}), designed to address high-dimensional model shifts in the presence of covariate shifts, thereby transferring knowledge from source tasks to the target task. The method consists of two steps. First, we perform a co-training step using both source and target samples, leveraging the $\ell_0$-sparsity of $\bz$ and the $\ell_1$-sparsity of the contrast $\bdk$.  We show that when the source tasks are sufficiently diverse, see Definition~\ref{def:diversity}, performing the first step of the TransFusion method suffices to ensure a fast rate. When such a condition is not met, we further perform a second step by fine-tuning the model on the target dataset. The method is shown to be rate-optimal and robust to covariate shifts. 

\subsection{Step 1: Co-Training} 
\label{jointtraining}
We start with the first step, a co-training step involving both target and source samples. {The challenge is tackling the distribution shifts while extracting the shared pattern between source and target samples to estimate $\bz$. Pooling all data as \emph{i.i.d.} samples leverages larger sample size and reduces noise, but suffers from large bias if the source and target distributions differ significantly. In contrast, training exclusively on the target sample prevents any information transfer from the source samples. It is therefore critical to to strike a balance between these two extremes to improve the estimation of $\bz$.} 
To this end, we propose a co-training step that estimates $\bk$s by solving the following problem:
\begin{multline}\label{obj}  
\hat{\bb} \in \argmin_{\bb \in \mathbb{R}^{(K+1)p}}   \left\{ \frac{1}{2N} \sum^{K}_{k=0} \|\byk-\bXk \bk\|_{2}^{2}\right.  \\
    \left. +\lambda_{0} \Big(\|\bz\|_{1} + \sum^{K}_{k=1}a_{k}\|\bk-\bz\|_{1} \Big) \right\},
\end{multline}
where $N=Kn_{S} + n_{T}$ is the total sample size, $\lambda_{0}$ is the tuning parameter and $\{a_{k}\}_{k = 1}^K$ are weights that will be specified later. In~\eqref{obj}, the first term measures the average fitness of the models with parameter $\{\bk\}_{k = 0}^K$, while the fused-regularization term simultaneously promotes the sparsity of $\bz$ and captures the $\ell_1$-sparse contrast between $\bz$ and $\bk$ by penalizing their  difference. 

We construct the first-step estimator as $\hat{\bw} = \frac{n_{S}}{N}\sum^{K}_{k=1} \hbk + \frac{n_{T}}{N}\hbz$. The motivation for this estimator is twofold: first, averaging across both target and source estimators utilizes the full sample, yielding an estimator with low variance. In addition, when the source datasets are sufficiently diverse, the bias of $\hat{\bw}$ is small. When the reduction in variance dominates the increase in bias, the one-step estimator $\hat{\bw}$ serves as a promising estimator of $\bz$ than using the target sample alone. 

We now formally characterize the source task diversity.
\begin{definition}[Source task diversity]\label{def:diversity}
Given $\bb \in \Theta(s, \bh)$, we quantify the diversity across source tasks with the metric 
$\|\frac{n_{S}}{N}\sum^{K}_{k=1} \bdk\|_{1} \leq \varepsilon_D$, where $\bdk: = \bk - \bz$ is the task-specific signal.
\end{definition}

A small $\varepsilon_{D}$ implies that  $\{\bk\}_{k = 1}^K$ are centered around $\bz$ and cover all the directions, such that the average parameter $\bw:= \frac{n_{S}}{N}\sum^{K}_{k=1}\bk + \frac{n_{T}}{N}\bz$ does not align with any direction significantly more than $\bz$. This kind of assumption is commonly imposed in transfer learning settings (\cite{du2020few},\cite{tripuraneni2020theory}).

We proceed to establish the statistical estimation rate for $\hat{\bw}$ under the setting of diverse source tasks.
\begin{theorem}
\label{stepone}
   Under Assumption \ref{A1} and \ref{A2}, if $n_{S} \gg s \log p$, then by choosing $\lambda_{0}=c_{0}\sqrt{\log p / N}$ for some universal constant $c_{0}$ and $a_{k}=8  \sqrt{n_{S}/N}$, we have
\begin{align}
\label{w_est_rate}
\|\hat{\bw}-\bw\|_{2}^2\lesssim \frac{s \log p}{N}+ (1+v_{n})\bar{h}\sqrt{\frac{\log p}{n_{S}}},
\end{align}
and 
\begin{align}
\label{steponeupperbound}
\|\hat{\bw}-\bz\|_{2}^2\lesssim \frac{s \log p}{N}+ (1+v_{n})\bar{h}\sqrt{\frac{\log p}{n_{S}}}+\varepsilon_{D}^2,
\end{align}
with probability at least $1-c_{1}\exp(-c_2 n_{T}) -c_{3}\exp \left(-c_{4} \log p \right)$, where $v_{n} := \sqrt{K^2 \log p / n_{S}}\bar{h}$ and $\bar{h}:=\frac{n_{S}}{N}\sum^{K}_{k=1} h_k$.
\end{theorem}

Let us break down the upper bound provided by equation~\eqref{steponeupperbound}. The first term, $s \log p/N$, represents the rate from estimating an $s$-sparse coefficient $\bz$ based on $N = K n_{S} + n_{T}$ i.i.d. samples. This term reveals the benefit of using both the source and target datasets for estimating the target parameter $\bz$. The second term, $\bar{h}\sqrt{\log p/n_{S}}$, accounts for the estimation error of $\bdk$ unique to each source task and thus is limited by the source sample size $n_S$. The factor $v_{n}$ is sample dependent and is negligible when $n_{S} \gg \bar{h}^2K^2\log p$. The first two terms together quantify the estimation error $\|\hat{\bw} - \bw\|_2^2$. The 
  third term, $\varepsilon_{D}^2$, measures the difference between $\bw$ and $\bz$ and contributes to the bias introduced by averaging. Notably, to obtain the bound (\ref{steponeupperbound}), we do not require a homogeneous distribution of the covariates $\bX^{(k)}$s but only impose the mild tail assumption as outlined in Assumption \ref{A1}, and the bound does not depend on the target sample size $n_{T}$.

As a comparison, if we apply the LASSO regression on the target data, the estimation error is of order $O(s \log p / n_{T})$. Therefore, if $N \gg n_{T}$, $n_{S} \gg \bar{h}^2 (n_{T}^2 \vee K^2 \log p)$ and $\varepsilon_{D} \ll \sqrt{s \log p / n_{T}}$, that is, the source tasks are sufficiently diverse with adequately large sample size, then one-step TransFusion method achieves a sharper estimation rate.  This corroborates our design intuition and quantitatively shows the benefit of transferring information from diverse source tasks even under covariate shifts.

\begin{remark}[Adaptive version of TransFusion]
    In Theorem \ref{stepone}, we choose a weight $a_k$ that does not depend on $h_k$, as we treat $h_k$ as an unknown priori. As a compromise, the estimation rate depends on $\bar{h}$, the averaged magnitude of model shifts.
    In fact, for a general choice of $\lz$ and $a_{k}$, TransFusion could yield a bound
    $$
     \|\hat{\bw} - \bz\|_{2}^2  \lesssim s\lz^2 + \sum^{K}_{k=1} a_{k}\lz h_{k} + \varepsilon_{D}^2
    $$    
    under certain conditions (cf. Lemma \ref{steponel1bound}). So if we have some information on $h_k$, we may adjust $a_k$ accordingly, focusing more on informative datasets with small $h_k$ and less or not at all on those with large $h_k$. In such cases, this adaptive version of TransFusion could potentially yield a fast estimation rate that is less sensitive to the magnitude of model shifts.
\end{remark}

\begin{remark}[Scalability with task number $K$]
   { TransFusion incorporates a novel fused regularizer  capturing the task-specific signals in the joint learning step. This technique robustifies the method against covariate-shift and introduces a dependency of the convergence rate on $K$ as a tradeoff. 
Specifically, the convergence rate of the first-step estimator is given by (\ref{w_est_rate}) with $v_{n}:= \sqrt{K^2 \log p / n_{S}}\bar{h}$ due to the non-strong convexity of the local empirical loss (cf. Lemma \ref{Thm1Lemma2}). If we increase $K$ while fixing  $n_{S}$, for large $K$, the sum will be dominated by the second term, which grows with $K$. 
Otherwise, if we increase $n_S$ with $K$, TransFusion would have a consistent error improvement.
This is supported by the simulation results and discussions in Appendix~\ref{app:suppl-sim}.}
\end{remark}

\subsection{Step 2: Local Debias}

Despite its merits, the one-step TransFusion method may experience large bias when  $\varepsilon_{D}$ is large. This is especially the case when the source tasks exhibit a skewed model shift towards one specific direction than $\bz$. In such cases, we employ an additional debias step that refines the initial estimator $\hat{\bw}$ and mitigates the impact of $\varepsilon_{D}$. Specifically, we correct $\hat{\bw}$ using the target sample as:
\begin{align}
\label{transferstep}
& \hbd \in \underset{\bd \in \mathbb{R}^p}{\operatorname{argmin}}
\left\{\frac{1}{2 n_{T}}\left\|\byz-\bXz \hat{\bw}-\bXz \boldsymbol{\delta}\right\|_{2}^{2}+\tilde{\lambda}\|\boldsymbol{\delta}\|_{1}\right\}, \nonumber \\
& \hbz_{\TransFusion}=\hat{\bw}+\hat{\boldsymbol{\delta}} .
\end{align}
Next, we demonstrate that with an appropriate choice of estimation strategy and tuning parameters, we can attain an optimal estimation rate of $\bz$ without requiring a small $\varepsilon_D$. Define the event 
\begin{align}\label{eq:def_A}
  A = \big\{s\log p / n_{S} \ge \bar{h} \sqrt{\log p /n_{T}} \big\},  
\end{align}
and $A^c$ as its complement. The following theorem establishes an upper bound on the estimation error for the two-step TransFusion algorithm.
\begin{theorem}
\label{steptwo}
Under the assumptions of Theorem 1, if 
$n_{T} \gtrsim s \log p$, $n_{S} \gtrsim  K^2 s \log p$ and $\bar{h}\sqrt{\log p / n_{T}} + Ks\log p/ n_{S}=o(1)$, then by choosing the parameters 
 \begin{align*}
      \lambda_{0} &= c_{0} \left(\sqrt{\frac{\log p}{N}} \mathbb{1}_{A} + \sqrt{\frac{\log p}{n_{S}}} \mathbb{1}_{A^c}\right) \text{, } \\
      a_{k} &=8 \left(\sqrt{\frac{n_{S
    }}{N}}\mathbb{1}_{A} + \frac{n_{S}}{N} \mathbb{1}_{A^c}\right),
  \end{align*}
  
and $\tilde{\lambda}=c_{1} \sqrt{\log p / n_{T}}$ for some universal constants $c_{0}$ and $c_{1}$, the solution of the two-step TransFusion method satisfies 
\begin{align}
\label{steptwoupperbound}
\|\hbz_{\TransFusion}-\bz\|_{2}^{2} &\lesssim \frac{s \log p}{N} + \bar{h}\sqrt{\frac{\log p}{n_{T}}},
\end{align}
with probability at least {$1 - c_{2} \exp \left(-c_{3} \log p \right)$.}
\end{theorem}

By comparing the results of Theorem~\ref{stepone} and~\ref{steptwo} [cf.~\eqref{steponeupperbound} and~\eqref{steptwoupperbound}], we see when 
$\sqrt{\log p/n_{T}} \lesssim \varepsilon_{D}^2/\bar{h}$, performing the second step improves the estimation precision. The ratio $\varepsilon_{D}/\bar{h}$ quantifies the normalized (by the magnitude of $\bdk$s) source task diversity, and thus our result shows applying the second step is beneficial for non-diverse source tasks. Note that the condition on the sample size $n_{T}$ and $n_S$ in Theorem~\ref{steptwo} is stronger than Theorem~\ref{stepone}. Such a condition is required to ensure the target-specific signals $\bdk$ being accurately captured to perform the correction in~\eqref{transferstep}. 




On the other hand, if $\sqrt{\log p/n_{T}} \gtrsim \varepsilon_{D}^2/\bar{h}$ applying the second step may even harm the model performance. Therefore, choosing between the one-step and two-step TransFusion methods carefully is key to getting the optimal estimation results. The following corollary provides guidelines for making this choice.

\begin{corollary}
\label{steptwooptimal}
     Under the assumptions of Theorem \ref{steptwo}, if we apply the one-step TransFusion method when $n_{T} \lesssim \log p / \bar{h}^2$ and apply the two-step TransFusion method otherwise, then obtained estimator {$\hbz_{\TransFusion-2}$} satisfies
    \begin{align}
    \label{comb_upperbound}
      \|\hbz_{\TransFusion-2}-\bz\|_{2}^{2} &\lesssim \frac{s \log p}{N} + \bar{h}\sqrt{\frac{\log p}{n_{T}}} \wedge \bar{h}^2 ,
    \end{align}
    with probability at least $1- c_{2} \exp \left(-c_{3} \log p \right)$.
\end{corollary}

Next, we establish the minimax optimality of the above strategy under certain conditions. The following result follows from minor modifications of Theorem 2 in \cite{li2022transfer}.

\begin{proposition}
\label{steptwolower}
  Under Assumption \ref{stepone} and  Assumption \ref{steptwo}, if {$N \gg s\log p$, $h_k \asymp \bar{h}$ and $\bar{h}\sqrt{\log p / n_{T}} = o(1)$}, then any estimator $\hat{\boldsymbol{\beta}}^{\prime}$ that is a measurable function of the sample $\{(\bXk, \byk)\}_{0\le k \le K}$ satisfies
  \small{
\begin{align}
    \inf _{\hat{\boldsymbol{\beta}}^{\prime}} \sup _{\boldsymbol{\beta} \in \Theta(s, \bh)}\left\|\hat{\boldsymbol{\beta}}^{\prime}-\boldsymbol{\beta}^{(0)}\right\|_2^2 \gtrsim \frac{s \log p}{N} + \frac{s\log p}{n_{T}} \wedge \bar{h}\sqrt{\frac{\log p}{n_{T}}} \wedge \bar{h}^2,
\end{align}
}

with probability at least $1/2$.
\end{proposition}

Comparing with the upper bound~\eqref{comb_upperbound}, we can conclude that, given the conditions outlined in Theorem~\ref{steptwo}, if source datasets are sufficient informative such that $\bar{h} \lesssim s \sqrt{\log p / n_{T}}$, then the proposed procedure is minimax optimal, even under covariate shifts.

\begin{remark}[Implementation of  TransFusion]
    Notice that although the two-step TransFusion method only involves one tuning parameter in each step, as discussed in Theorem \ref{steptwo} and Corollary \ref{steptwooptimal}, it relies on a dichotomous strategy that depends on the value of $\bar{h}$. However, it can still be practically applied without knowing $\bar{h}$ in advance by implementing both choices and selecting the one with smaller validation error. On the computation front, the global minimizer of each TransFusion step can be efficiently found by numerical algorithms such as iterative soft-thresholding. The implementation details are provided in Appendix~\ref{app:TransFusion-implementation}.
\end{remark}

\subsection{Understanding the Robustness of the TransFusion Method to Covariate Shift}
\label{subsec:discussion}
In this section, we discuss the underlying mechanisms that make the TransFusion method robust to covariate shifts via comparing with the two-step method proposed in \cite{li2022transfer} and \cite{tian2022transfer}. While both methods aim to address the high-dimensional transfer learning problem, we take a significantly different approach in the first co-training step.

In their approach, the first step pools samples from both target and source tasks, and performs a sparse regression to obtain the initial estimator. In the linear regression setting, this estimator comes with an asymptotic bias expressed as
\begin{align}
\label{Translassobias}
{\bd_{\text{Pooling}}}:=\left(\sum_{k=1}^K \Sigk\right)^{-1} \sum_{k=1}^K \Sigk \bd^{(k)}.
\end{align}
Due to  the weights introduced by the  covariance matrices,  the contrast $\bdk$s can be amplified by a factor of  $C_{\Sigma}$, defined as {\small
$$
C_{\Sigma}:=1+\max _{j \leq p} \max _{k}\left\|e_j^{\top}\left(\Sigk-\Sigz\right)\left(\sum_{1 \le k \le K}\frac{1}{K}\Sigk\right)^{-1}\right\|_1.
$$
}

Consequently, in the linear regression setting, their estimator yields the following estimation rate~\cite[Theorem 4]{li2022transfer}:
$$
\frac{s \log p}{N} + \left(C_{\Sigma} \sqrt{\frac{\log p}{n_{T}}}\bar{h} \right) \wedge\left(C_{\Sigma}^2 \bar{h}^2\right).
$$
When the $\Sigk$s are dissimilar, the factor $C_{\Sigma}$ can diverge with dimension $p$ even if Assumption~\ref{A1} holds, considerably deteriorating the estimation accuracy. See Appendix~\ref{app:c_sigma} for a detailed discussion. 

In contrast, as shown in (\ref{comb_upperbound}), our method is robust to such covariance heterogeneity and thus doesn't involve the $C_{\Sigma}$ factor. This is achieved by incorporating a fused-regularizer, allowing us to accurately capture task-specific signals under covariate shifts. Solving the  objective leads to an initial estimator with asymptotic bias
$$
\bd_{\TransFusion} := \frac{1}{K}\sum_{k=1}^K \bdk,
$$
which is free from the impact of the covariance matrices. 
This bias is much smaller than $\bd_{\text{Pooling}}$ under covariate shift settings with a large $C_\Sigma$. 

\section{D-TranFusion: Distributed Transfer Learning in One-Shot}
\label{sec:D-TransFusion}
In this section, we consider the distributed transfer learning problem where the target and $K$ source samples are stored by different computing nodes. Such a setting is of primary interest in learning problems involving a massive amount of training data, where brute-forcely pooling the raw data is not admissible due to practical constraints such as storage limitation, communication cost, and privacy concerns. 

This motivates us to consider developing a communication-efficient distributed TransFusion method, termed D-TransFusion. Our method is based on  TransFusion  and leverages the idea of divide-and-conquer to facilitate communication efficiency, aiming to achieve a comparable estimation error as  TransFusion but using only one-shot communication. Specifically, D-TransFusion consists of the following two steps.

\textbf{Step 1.} Each node $k$ computes an estimator $\tbk$ (to be specified later) locally based on source sample $(\bXk,\byk)$ and transmits it to the target node. The target node then 
aggregates them with its own sample $(\bXz,\byz)$ via solving the following  
problem:\vspace{-0.2cm}
\begin{multline}\label{com_eff_obj}
  \hat{\bb}_C \in \argmin_{\bb \in \mathbb{R}^{(K+1)p}}~\bigg\{ \frac{1}{2N} \sum^{K}_{k=1} \|\sqrt{n_S}(\tbk-\bk)\|_{2}^{2} \\
  + \frac{1}{2N}\|\byz-\bXz \bz\|_{2}^{2} 
  +\lambda_{0} \mR(\bb) \bigg\},
\end{multline} 
where $\mR(\bb):=\|\bz\|_{1} + \sum_{k=1}^{K} a_{k}\|\bk-\bz\|_{1}$. With the solution $\hat{\bb}_C $, the target node computes $\hat{\bw}_{C} = \frac{n_{S}}{N}\sum^{K}_{k=1} \hbk_{C} + \frac{n_{T}}{N}\hbz_{C}$.

\textbf{Step 2.} The target node corrects $\hat{\bw}_{C}$ on its local sample $(\bXz,\byz)$ by solving
{\small
\begin{align*}
    \hbd_{C} \in \underset{\boldsymbol{\delta} \in \mathbb{R}^p}{\operatorname{argmin}}
\left\{\frac{1}{2 n_{T}}\left\|\byz-\bXz \hat{\bw}_{C}-\bXz \boldsymbol{\delta}\right\|_{2}^{2}+\tilde{\lambda}\|\boldsymbol{\delta}\|_{1}\right\},
\end{align*}}
and outputs the estimator $\hbz_{\DTransFusion}=\hat{\bw}_{C}+\hat{\boldsymbol{\delta}}_{C}$.

Comparing with~\eqref{obj} and~\eqref{transferstep}, we can see that D-TransFusion differs from the centralized TransFusion method only in the first step.
To avoid the involvement of source samples, D-TransFusion replaces the least square loss $\|\byk - \bXk \bk\|_2^2$  by the squared loss $\|\sqrt{n_S}(\tbk - \bk)\|_2^2$, wherein $\tbk$ serves as a ``pseudo sample'' summarizing the information of $\bk$ that the $k$-th source sample contains. By doing so, only one-shot communication is required to transmit the summary statistics $\tbk$ from the source to the target node, significantly reducing the communication overhead. Here, $\tbk$ is carefully selected as a de-biased LASSO estimator \citep{javanmard2014confidence} that minimizes estimator variance while also controlling the bias under a given threshold. See Appendix \ref{DebiasEst} for a detailed discussion. This choice ensures that D-TransFusion can significantly reduce communication overhead while achieving the
minimum loss of sample efficiency compared to centralized TransFusion. 

More importantly, D-TransFusion  allows for pre-training on each source data nodes before transfer learning. The decoupling of training on the source and target samples eliminates the need for training from scratch when the target samples change,  and thus enhances the model's adaptability to downstream tasks.

We now establish the statistical precision of the one-step D-TransFusion method. Define $\delta_k = \frac{s \log p}{N} + \frac{n_{S}}{N} \sqrt{\frac{\log p}{n_{S}}}h_{k}$ and $\delta_0 =\frac{Ks \log p}{N} +\sqrt{\frac{\log p}{n_{S}}}\bar{h}$. The following theorem provides an upper bound for the estimation error of the one-step D-TransFusion estimator $\hat{\bw}_{C}$.
\begin{theorem}
    \label{ComTransFusion}
Under Assumption \ref{A1} and \ref{A2} and the assumptions $n_{S} \gg Ks^2\log p$, $n_{S} \gtrsim (\bar{h}^2\vee K^2)s \log p$, $h_{k} \asymp \bar{h}$, if we construct $\{\tilde{\bb}^{(k)}\}_{k = 1}^K$ through (\ref{Comobj}) and (\ref{JMalgorithm}) with parameters $\tilde{\lambda}_k = \mu_{k} =c_{1}\sqrt{\log p / n_{S}}$, and solve problem (\ref{com_eff_obj}) with parameters $\lambda_{0}$ and $\{a_{k}\}_{k=1,\dots,K}$ chosen such that
\begin{align}
\label{para_requirement}
 \lambda_{0}&=c_{0} \left(\sqrt{\frac{\log p}{N}} + \delta_{0} \right), \nonumber\\ 
 a_{k}\lz&=c_{0}\Big(8 \vee \frac{\bar{h}}{h_{k}} \Big) \! \left(\sqrt{\frac{n_{S}}{N} \frac{\log p}{N}}+\delta_{k} \right),
\end{align}
for some universal constant $c_{0}$ and $c_1$, then with probability at least $1- c_{2}\exp \left(-c_{3}n_{T} \right) - c_{4} \exp \left(-c_{5} \log p \right)$,
{\small
\begin{align}
\label{D-TransFusionbound1}
\|\hat{\bw}_{C}-\bz\|_{2}^2\lesssim s{\frac{ \log p}{N}}+ \sqrt{\frac{\log p}{n_S}}\bar{h} + \varepsilon_{D}^2+ s\delta_{0}^2 + \sum^{K}_{k=1} \delta_k  h_{k},
\end{align}
}
if we further assume $\bar{h} = O(1)$, then we have
\begin{align}
\label{D-TransFusionbound2}
\|\hat{\bw}_{C}-\bz\|_{2}^2\lesssim s{\frac{ \log p}{N}}+ \sqrt{\frac{\log p}{n_S}}\bar{h} + \varepsilon_{D}^2.
\end{align}
\end{theorem} 

Let us now compare the result to Theorem \ref{stepone}. Under the additional assumption $n_{S} \gg Ks^2\log p$, $n_{S} \gtrsim (\bar{h}^2\vee K^2)s \log p$, $h_{k} \asymp \bar{h}$, which requires the source tasks roughly equally informative with sufficiently large size, the estimation error of D-TransFusion is larger than TransFusion by $s\delta_{0}^2 + \sum^{K}_{k=1} \delta_k  h_{k}$. This reflects the cost of sample efficiency for achieving one-shot communication. When $\bar{h} = O(1)$, such difference is negligible compared to the estimation error of TransFusion, the statistical accuracy of $\hat{\bw}_{C}$ matches that of the centralized counterpart $\hat{\bw}$ in the asymptotic sense. 

The second step of D-TransFusion can be designed in analogue to that of TransFusion. 
Recall the event $A = \{s\log p / n_{S} \ge \bar{h}/ \sqrt{\log p /n_{T}} \}$ defined in~\eqref{eq:def_A}. The following theorem establishes the statistical estimation rate of the final estimator obtained from the two-step D-TransFusion method.

\begin{theorem}
\label{ComStep2}
    Under the assumptions of Theorem \ref{ComTransFusion}, if further assume $n_{T} \gtrsim s \log p$, $\bar{h}\sqrt{\log p / n_{T}}=o(1)$, then by choosing $\lambda_{0}$ and $\{a_{k}\}_{k=1,\dots,K}$ such that
\begin{align*}
\label{para_requirement2}
\lz&=c_{0} \left(\sqrt{\frac{\log p}{N}} \mathbb{1}_{A} + \sqrt{\frac{\log p}{n_{S}}} \mathbb{1}_{A^c} + \delta_{0} \right),\\
a_{k}\lz&=c_{0} \Big(8 \vee \frac{\bar{h}}{h_{k}} \Big)  \left(\sqrt{\frac{n_{S
    }}{N} \frac{\log p}{N}}+\delta_{k} \right), 
\end{align*}
and
$\tilde{\lambda}=c_{1} \sqrt{\log p / n_{T}}$ for some universal constants $c_{0}$ and $c_{1}$, we have
\begin{align}
\|\hat{\bb}^{(0)}_{\DTransFusion}-\bb^{(0)}\|_{2}^{2} \lesssim \frac{s {\log p}}{N}+\sqrt{\frac{{\log p}}{n_{T}}} \bar{h},
\end{align}
with probability at least {$1- c_{2} \exp \left(-c_{3} \log p \right)$}.
\end{theorem}
Theorem \ref{ComStep2} ensures that the two-step D-TransFusion method achieves a statistical rate of the same order as the centralized two-step TransFusion method {under the previously discussed additional conditions}. By employing similar reasoning as in Corollary \ref{steptwooptimal} and Proposition \ref{steptwolower}, we can further establish conditions under which D-TransFusion is minimax optimal. These results demonstrate that D-TransFusion is an efficient and robust solution when dealing with large-scale distributed datasets with covariate shifts.

\begin{remark}[The efficacy of D-TransFusion]
{D-TransFusion aims to address the scenario where the source datasets are not co-located and cannot to be merged. This differs from the traditional distributed computing paradigm, where one splits the whole data into parts and parallelizes the cost due to the large data size. 
As for the implementation cost,  since  $K$ debiased lasso estimators are computed in step 1 and transmitted in step 2, it requires per source node storing and transmitting a $p$-dimensional vector. The computation cost readily follows that of the debiased and standard lasso, provided in \cite{lee2017communication}. Although concerns may arise about the computation cost of the debiased Lasso, it is noteworthy that under mild conditions, the de-biased lasso estimator $\tbk$ can be replaced by other asymptotically unbiased estimator such as the SCAD estimator (\cite{fan2001variable}), which enables the D-TransFusion to enjoy both comparable computational complexity, but distributed to $K$ parallel processors, and statistical precision to its non-distributed counterpart given a moderate task number $K$.}
\end{remark}

\section{Simulation}
\label{sec:simulation}
We evaluate the empirical performance of our proposed methods, \textit{TransFusion}  and \textit{D-TransFusion}, and compare with existing methods including \textit{Trans-Lasso} \citep{li2022transfer} and \textit{TransHDGLM} \citep{li2023estimation}. For two-step approaches, we report and compare the performance of both steps to better understand scenarios where the second step is necessary. As a baseline, we include the estimation error obtained by the LASSO regression on the target task, which we call \textit{Lasso (baseline)}.
Each simulation setting is replicated with 100 independent trials, and we report the average performance. All methods are implemented based on R package \textit{glmnet} with standard configuration, and parameters are chosen via 10-fold cross-validation. 
\\[1ex]
We follow a similar experimental setup as in \cite{li2022transfer} and \cite{li2023estimation} by considering a high-dimensional linear regression problem with  $p = 500$ and sparsity level $s = 10$. The target model  is set as $\bz_{j} = 0.3$ for $1 \le j \le s$ and $\bz_{j} = 0$ otherwise.
We generate $n_{T} = 150$ independent target samples $(\bXz, \byz)$  by $\byz = \bXz \bz + \bez$ with $\bXz_{i \cdot} \sim N(0, \boldsymbol{I})$ and $\bez_{i} \sim N(0, 1)$. 

The source sample size is set to be $n_{S} = 200$, and the source task number $K$ varies in the range $\{1,3,5,7,9\}$. We set $h_k = 12$ for $1 \le k \le K$. To simulate model and covariate shift we consider the following parameter configurations for the source tasks.\\[1ex]
\noindent\textbf{Model Shift.} To investigate the impact of task diversity, we
simulate two types of model shifts.\\
\emph{ \textbf{(i)} Diverse source tasks.} For $k = 1, \ldots, K-1$ we set  $\bk = \bz + \bdk$ with $\bdk_{j} \sim N(0, (h_{k}/50)^2)$ for $1 \le j \le 50$ and $\bdk_{j}=0$ otherwise. The last source model is generated with $\bd^{(K)} = -\sum^{K-1}_{k=1} \bdk$ so that the task diversity measure $\varepsilon_D = 0$.  \\[0.5ex]
{\it \textbf{(ii)} Non-diverse source tasks.} Each $k$-th task-specific signal is generated as $\bdk_{j} \sim N(0.1, (h_{k}/50)^2)$ for $1 \le j \le 50$ and $\bdk_{j}=0$ otherwise.\\[1.5ex]
\noindent\textbf{Covariate Shift.}
To demonstrate the robustness of TransFusion  to covariate shifts, we consider two settings with different covariate distributions. In each setting, we generate $n_{S}$ independent samples for each source task.\\[0.5ex]
{\it \textbf{(a)} Homogeneous design.}
Each  $\bXk_{i \cdot} \sim N(0, \boldsymbol{I})$. \\[0.5ex]
{\it \textbf{(b)} Heterogeneous design.}
Each $\bXk_{i \cdot} \sim N(0, \Sigk)$ with $\Sigk = (\boldsymbol{A}^{(k)})^\top (\boldsymbol{A}^{(k)}) + \boldsymbol{I}$. Here $\boldsymbol{A}^{(k)}$ is a random matrix with each entry equals 0.3 with probability 0.3 and equals 0 with probability 0.7.

\begin{figure}
  \centering
  \includegraphics[width = 0.5\textwidth]{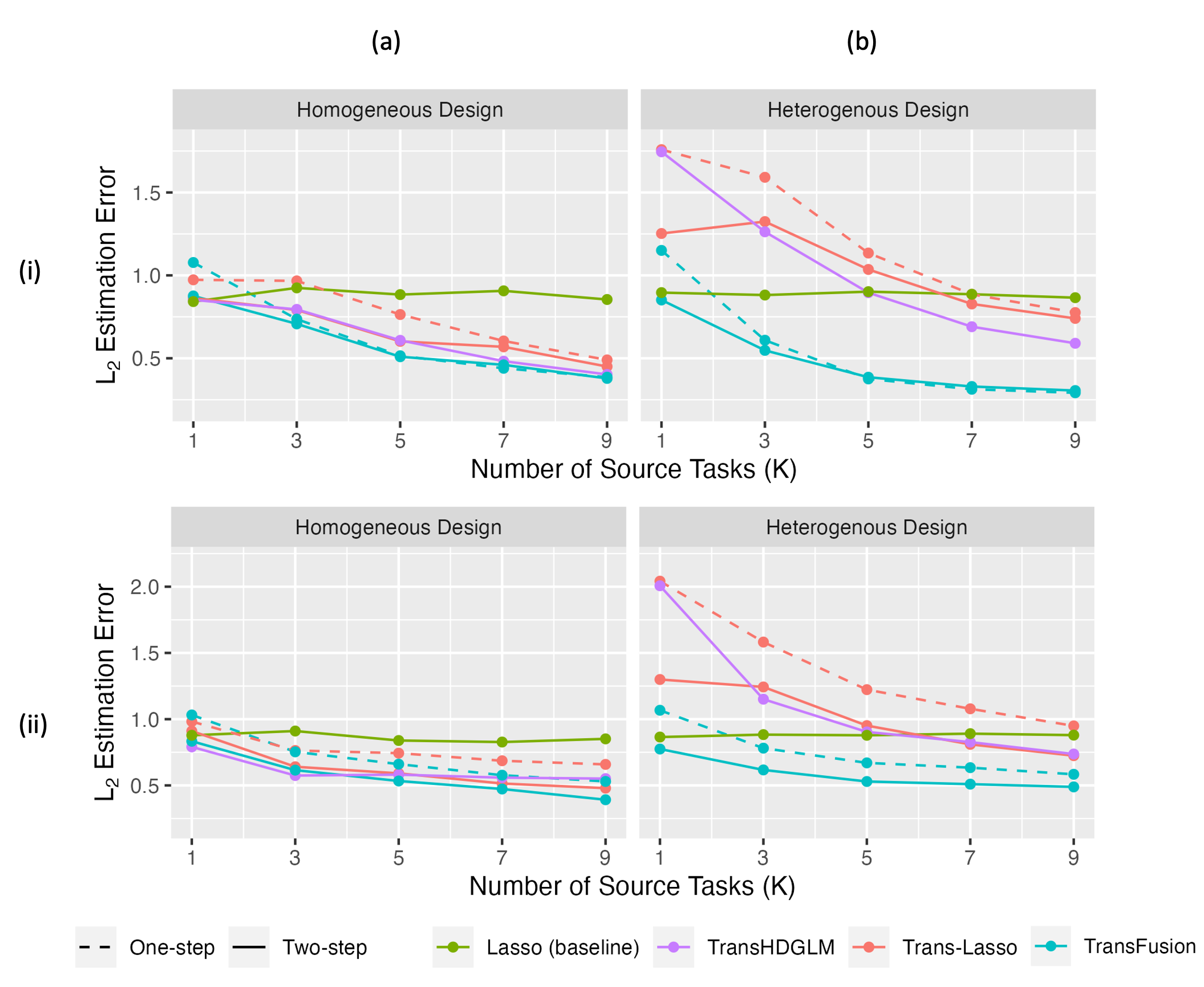}\vspace{-0.3cm}
\caption{Comparison of estimation errors under (i) diverse and (ii) non-diverse source task settings with (a) homogeneous design and (b) heterogeneous design.}
  \label{fig:four}
\end{figure}

\begin{figure}
  \vspace{-.2in}
  \centering
  \includegraphics[width = 0.45\textwidth]{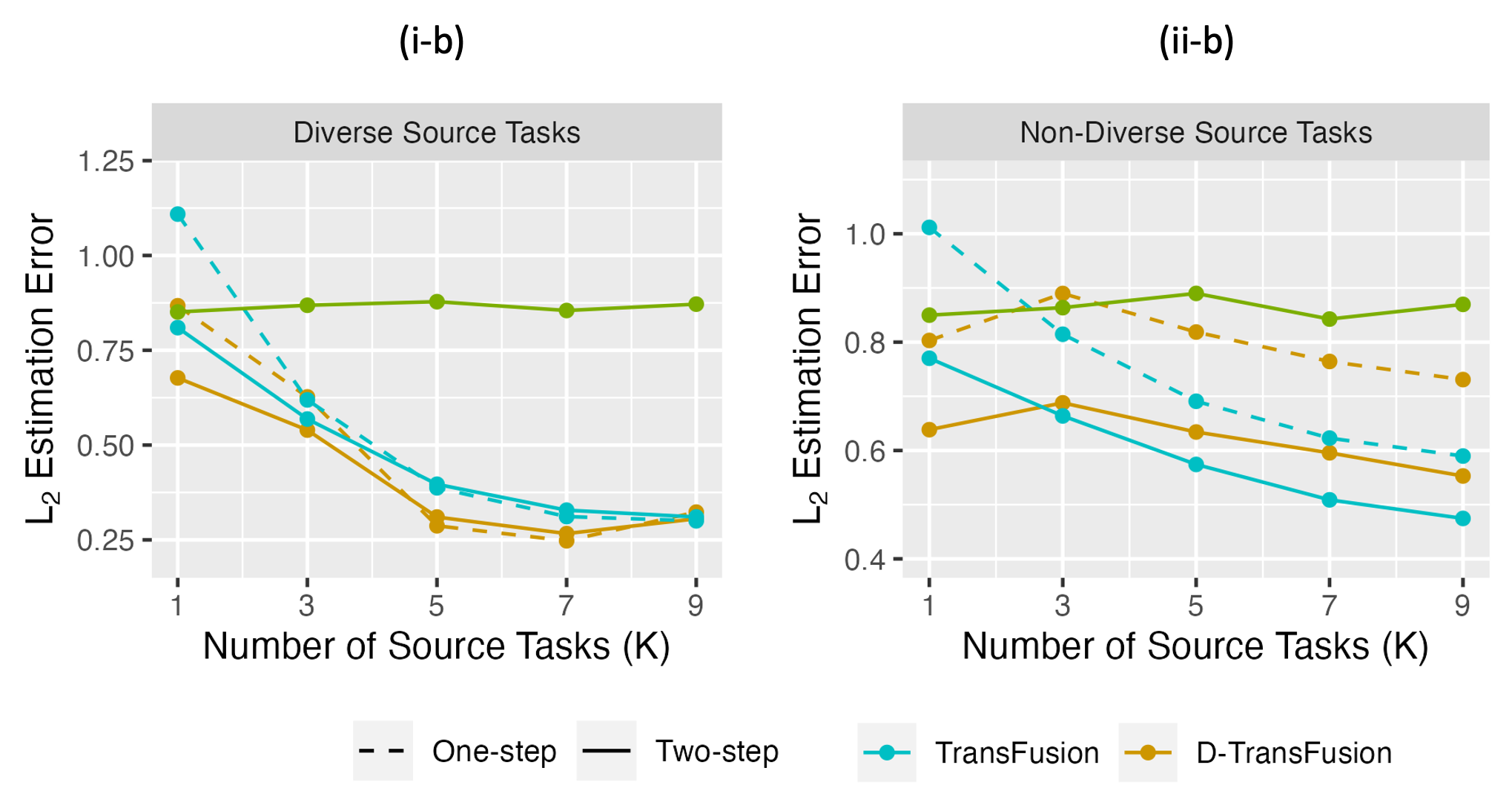}\vspace{-0.3cm}
\caption{Comparison of the estimation errors of D-TransFusion and TransFusion methods under (i) diverse and (ii) non-diverse source task settings with (b) heterogeneous design.}
  \label{fig:two}
 \vspace{-.05in}
\end{figure}

We consider four experimental settings based on combinations of model design (i) and (ii) with covariate design (a) and (b) to generate the source samples.  Fig.~\ref{fig:four} reports the $\ell_2$ estimation error of $\bz$ versus the source task number $K$ in these four settings. Fig.~\ref{fig:two} reports a focused comparison between the performance of D-TransFusion and TransFusion in heterogeneous design (b). {More simulation results are reported in the Appendix \ref{simulation}.} 
The following comments are in order.\\[1ex]

$\bullet$ Robustness to covariate shift. Fig.\ref{fig:four} (i-a) 
shows in the diverse source tasks setting (i) when there is no covariate shift {(a setting in favor of {\it Trans-Lasso})}, TransFusion achieves comparable estimation error with the state-of-the-art methods. However, when covariate shift exists, a comparison of Fig.\ref{fig:four} (i-b) with (i-a) shows the estimator errors obtained by both {\it Trans-Lasso} and {\it TransHDGLM} increase significantly, and are even larger than {\it Lasso (baseline)} for small values of $K$. On the contrary, the performance of TransFusion remains the same, showing its robustness to covariate shift. When the source tasks are non-diverse, Fig.\ref{fig:four} (ii-a) and (ii-b) reveal similar advantages achieved by two-step TransFusion. Consequently, TransFusion offers better reliability when the degree of covariate shift and source task diversity are unknown a priori.

$\bullet$ Impact of source task diversity. In Theorem~\ref{stepone} we have proved that when the source tasks are sufficiently diverse, applying one-step TransFusion suffices to obtain a small estimation error. Fig.\ref{fig:four} (i-a) and (i-b) corroborate this statement: under both homogeneous and heterogeneous covariate designs, one- and two-step TransFusion yields comparable estimation errors. In the more challenging setting with non-diverse source tasks, Fig.\ref{fig:four} (ii-a) and (ii-b) indicate applying the second de-bias step  reduces the estimation error, which is also consistent with the implications of Theorem~\ref{steptwo}. 

 
$\bullet$ D-TransFusion matches the performance of TransFusion. Fig.~\ref{fig:two} demonstrates for small $K$, D-TransFusion attains comparable or even smaller estimation error than TransFusion, but uses only one-shot communication and has the ability to quickly adapt to downstream tasks. As $K$ increases, the gap reduces, and TransFusion gradually outperforms D-TransFusion. This is because D-TransFusion has a more restrictive growth condition on the source sample size $n_S$ with $K$, a requirement common to divide-and-conquer type methods. It also reveals the tradeoff between sample efficiency and communication cost.

{In addition to the simulation study using synthetic data, the covariate-shift robustness of \textit{TransFusion} method is further validated on real-world application through the MNIST handwritten digit classification task. The results are provided in Appendix \ref{app:real-data}.}

\section{Conclusion}
\label{sec:conclusion}

{In this paper, we introduce a novel solution to tackle model shifts in high-dimensional transfer learning problems, ensuring robustness to covariate shifts and efficiency in knowledge transfer across distributed datasets. We provide a theoretical guarantee, showing its capacity to fully utilize the source samples to achieve an optimal estimation rate with one-shot communication and under covariate shifts. Simulation results validate our theory and showcase the state-of-the-art performance of the proposed method under various settings.}


\newpage
\bibliographystyle{plainnat}
\bibliography{bibfile}

\appendix
\thispagestyle{empty}

\onecolumn 

\section{Proof of Theorems}
\label{theoremproof}
In this section, we provide proof of all the theorems. Throughout the section, we adopt the following notations to analyze the solution of the problem (\ref{obj}):
\begin{align}
\label{transrule}
\by:=\left(\begin{array}{c}
\by^{(1)} \\
\by^{(2)} \\
\vdots \\
\by^{(K)} \\
\by^{(0)}
\end{array}\right) \quad  \bX:=\left(\begin{array}{ccccc}
\bX^{(1)} & 0 & \cdots & 0 & \bX^{(1)} \\
0 & \bX^{(2)} & \cdots & 0 & \bX^{(2)} \\
\vdots & \vdots & \vdots & \vdots & \vdots \\
0 & 0 & \cdots & \bX^{(K)} & \bX^{(K)} \\
0 & 0 & \cdots & 0 & \bX^{(0)}
\end{array}\right) \quad \btheta^{*}= \left(\begin{array}{c}
(\btheta^{*})^{(1)} \\
(\btheta^{*})^{(2)} \\
\vdots \\
(\btheta^{*})^{(K)} \\
(\btheta^{*})^{(0)}
\end{array}\right) := 
\left(\begin{array}{c}
\bb^{(1)}-\bb^{(0)} \\
\bb^{(2)}-\bb^{(0)} \\
\vdots \\
\bb^{(K)}-\bb^{(0)} \\
\bb^{(0)}
\end{array}\right).
\end{align}
where recall that $\bz$ is the target model parameter and $\bk$s are source model parameters.


Under this transformation, solving problem (\ref{obj}) is equivalent as solving
\begin{equation}
\label{transobj}
\hat{\btheta} = \underset{\btheta}{\operatorname{argmin}} \{\mL(\btheta)+\lambda_{0} \mR(\btheta)\},
\end{equation}
where we define $\mL(\btheta) := \frac{1}{2N}\left\|\by-\bX\btheta\right\|_{2}^{2}$ and $\lambda_{0}\mR(\btheta) := \lambda_{0} \left\| \btheta^{(0)} \right\|_{1} + \lambda_{0}\sum^{K}_{k=1} a_{k} \left\| \btheta^{(k)} \right\|_{1} = \sum^{K}_{k=0} \lambda_{0}a_{k}\left\| \btheta^{(k)} \right\|_{1}$ for any $\btheta \in \mathbb{R}^{(K+1)p}$. {Since there exists a one-to-one transformation between $\btheta^{*}$ and $\bb$, we can quantify the estimation error $\hat{\bb} - \bb$ by analyzing $\hat{\btheta} - \btheta^{*}$.}

We first establish an essential property of sub-Gaussian design matrices, which serves as fundamental building blocks for our subsequent analysis. The following lemma \ref{RSC} shows that the least square objective function has a \textit{restricted} strongly convex (RSC) and \textit{restricted} smooth (RSM) property. For a detailed discussion and proof of this lemma, interested readers can refer to Lemma 13 of \cite{loh2011high} and Lemma 6 of \cite{agarwal2010fast}. 
\begin{lemma}[RSC and RSM property] 
\label{RSC}
Under Assumption \ref{A1}, for any $\boldsymbol{\bDelta} \in \mathbb{R}^{p}$, with probability at least $1 - c_1 \exp (-c_2 n_k)$,
    \begin{align*}
    \frac{1}{n_k}\left\|\bX^{(k)}\boldsymbol{\bDelta} \right\|_2^2
&=
\boldsymbol{\bDelta}^{\top}\hat{\boldsymbol{\Sigma}}^{(k)}\boldsymbol{\bDelta} \geq \alpha_k\|\boldsymbol{\bDelta}\|_2^2-\beta_k \frac{\log p}{n_k}\|\boldsymbol{\bDelta}\|_1^2, \\
\frac{1}{n_k}\left\|\bX^{(k)}\boldsymbol{\bDelta} \right\|_2^2
&=
\boldsymbol{\bDelta}^{\top}\hat{\boldsymbol{\Sigma}}^{(k)}\boldsymbol{\bDelta} \leq \gamma_k\|\boldsymbol{\bDelta}\|_2^2+\tau_k \frac{\log p}{n_k}\|\boldsymbol{\bDelta}\|_1^2, 
    \end{align*}
    where $\alpha_{k} = \frac{1}{2}\Lambda_{\min}(\Sigk) \geq 1/c$,$\gamma_{k} = 2\Lambda_{\max}(\Sigk) \leq c$ and $\beta_{k}, \tau_{k} \leq c$, $n_{S} = n_{S}$ for $k = 1, \dots, K$ and $n_{T}$ for $k = 0$.
\end{lemma}

\subsection{Proof of Theorem \ref{stepone}}
 
Define $\hat{\bDelta}:=\hat{\btheta}-\btheta^{*}$ as the estimation error of $\hat{\btheta}$ and the corresponding k-th block {$\hat{\bDelta}^{(k)}:=\hat{\btheta}^{(k)}-(\btheta^{*})^{(k)}$}. {For brevity, we will omit the superscript and write $(\btheta^{*})^{(k)}$ as $\btheta^{(k)}$ for $0 \le k \le K$ when there is no ambiguity. }

Further define $\wev:=\sum_{k=1}^K \frac{n_S}{N} \hat{\bDelta}^{(k)}+\hat{\bDelta}^{(0)} = \hat{\bw}-\bw$ as the estimation error of the parameter average $\bw$. Our goal is to establish an upper bound for $\|\wev\|_2^2$.

The proof of Theorem \ref{stepone} relies on three key technical lemmas. The proof of these lemmas is in Appendix \ref{lemmaproof}.
The first lemma establishes an upper bound for the first-order term of the Taylor series expansion of $\mL(\btheta)$. 

\begin{lemma}
\label{concentration}
Under Assumption \ref{A1} and \ref{A2}, if $n_{S} \gtrsim \log p$, then by choosing $\lambda_{0}=c_{0} \sqrt{\frac{\log p}{N}}$ and $\lk=a_{k}\lz=c_{0} \sqrt{\frac{n_{S}}{N} \frac{\log p}{N}}$ for some appropriate constant $c_{0}$, we have for any $\bDelta = \left(\left(\bDelta^{(1)}\right)^{\top}, \dots, \left(\bDelta^{(K)}\right)^{\top}, \left(\bDelta^{(0)}\right)^{\top} \right)^{\top}\in \mathbb{R}^{(K+1) p}$,
$$
\left|\left\langle\nabla \mL\left(\btheta^{*}\right), \bDelta\right\rangle\right|
\leq \sum_{k=1}^{K} \frac{\lambda_{k}}{2}\left\|\bDelta^{(k)}\right\|_{1}+\frac{\lambda_{0}}{2}\left\|\bDelta^{(0)}\right\|_{1} .
$$

with probability larger than $1-c_{1} \exp \left(-c_{2} \log p \right)$.   
\end{lemma}

Recall that we define $\lk=a_{k}\lz$.  The next lemma establishes a restricted set of directions in which $\wev = \hat{\bw} - \bw$ lies.

\begin{lemma}
    \label{Thm1Lemma1}
     Under Assumption \ref{A1} and \ref{A2}, and the conditions of Lemma \ref{concentration}, if further assume $
\lk \geq  8\lambda_0 \frac{n_S}{N}
$ and $n_{S} > n_{T}$, then the estimation error $\wev$ satisfies the inequality
    \begin{align*}
        \sum_{k=1}^K \lambda_k\left\|\hat{\bDelta}^{(k)}\right\|_1+2\lambda_0\left\|\hat{\bDelta}^{(0)}\right\|_1 \leq 8 \lambda_0\left\|\wev_S\right\|_1+8 \sum_{k=1}^K \lambda_k h_{k},
    \end{align*}
with probability larger than $1 - c_{1}\exp (-c_{2}\log p)$, where $S$ is the support set of $\bb^{(0)}$.
\end{lemma}

The following lemma ensures a property analogous to restricted strong convexity for $\wev$.
\begin{lemma}    \label{Thm1Lemma2}
      Under Assumption \ref{A1} and \ref{A2} and the conditions of Lemma \ref{Thm1Lemma1},  the estimation error $\wev$ satisfies
\begin{align}
\label{WRSC}
\hat{\bDelta}^{\top}\hat{\bSig}\hat{\bDelta} 
 = \mL\left(\btheta^*+\hat{\bDelta}\right)-\mL\left(\btheta^*\right)-\left\langle\nabla \mL\left(\btheta^*\right), \hat{\bDelta}\right\rangle \geq (1 - u_{n})\alpha_{\min} \left\| \wev\right\|_2^2-v_{n} \sum_{k=1}^K \lambda_k h_{k} 
\end{align}
with probability larger than $1 - c_{1}\exp(c_2 n_{T}) - c_{3} \exp(c_4 \log p)$, where $\hat{\bSig} := \frac{1}{N} \bX^{\top}\bX$,
\begin{align*}
    u_{n}:=\frac{ 256\beta_{\max }\lz^2}{\alpha_{\min}\lambda^2_{k}\wedge(\lambda^2_{0}/(K+1)) }\frac{s \log p}{N} &, \quad v_{n} := \frac{256 \beta_{\max}}{\lambda^2_{k}\wedge(\lambda^2_{0}/(K+1)) } \frac{\log p}{N} \left(\sum_{k=1}^{K}\lk h_{k}\right), \\
    \alpha_{\min} := \min_{0 \le k \le K} \alpha_{k}&, \quad \beta_{\max} := \max_{0 \le k \le K} \beta_{k},
\end{align*}
with RSC constants $(\alpha_{k}, \beta_{k})$ defined in Lemma \ref{RSC}.
\end{lemma}

We now turn to the proof of the theorem. In the following proof, we make use of the function $F: \mathbb{R}^{(K+1)p} \rightarrow \mathbb{R}$, given by $$F(\bDelta)=\mL\left(\btheta^{*}+\bDelta\right)-\mL\left(\btheta^{*}\right)+\lambda_{0}\mR \left(\btheta^{*}+\bDelta\right)-\lambda_{0}\mR \left(\btheta^{*}\right),$$
 where $\btheta^{*}$ is the transformed model parameter defined in (\ref{transrule}), and $\bDelta=\left(\left(\bDelta^{(1)}\right)^{\top}, \ldots, \left(\bDelta^{(K)}\right)^{\top}, \left(\bDelta^{(0)}\right)^{\top}\right)^{\top} \in \mathbb{R}^{(K+1)p}$.

By Lemma \ref{concentration}, triangle inequality, and the fact that $\left\|\btheta^{(0)}_{S^c}\right\|_{1} = 0$,  $\left\|\btheta^{(k)}\right\|_{1} \le h_k$ for $1 \le k \le K$, we have
$$
\begin{aligned}
F(\hat{\bDelta})= & \mL\left(\btheta^{*}+\hat{\bDelta}\right)-\mL\left(\btheta^{*}\right)+\lambda_{0}\mR \left(\btheta^{*}+\hat{\bDelta}\right)-\lambda_{0}\mR \left(\btheta^{*}\right) \\
\geq & -\left\|\mL\left(\btheta^{*}\right)\right\|_{\infty} \| \hat{\bDelta} \|_{1}+ \hat{\bDelta}^{\top} \nabla^2 \mL\left(\btheta^*+\gamma \hat{\bDelta}\right) \hat{\bDelta} \quad(\gamma \in(0,1))  \\
& +\sum_{k=1}^{K}\lk\left(\left\|\btheta^{(k)}+\hat{\bDelta}^{(k)}\right\|_{1}-\left\|\btheta^{(k)}\right\|_{1}\right) + \lambda_{0}\left\|\btheta^{(0)} + \hat{\bDelta}^{(0)}\right\|_1 - \lambda_{0}\left\|\btheta^{(0)}\right\|_1 \\
\geq & -\sum_{k=1}^{K} \frac{\lambda_{k}}{2}\left\|\hat{\bDelta}^{(k)}\right\|_{1}-\frac{\lambda_{0}}{2}\left\|\hat{\bDelta}^{(0)}\right\|_{1}+\hat{\bDelta}^\top \hat{\bSig} \hat{\bDelta} \\
& +\sum_{k=1}^{K}\lk\left(\left\|\hat{\bDelta}^{(k)}\right\|_{1}-2\left\|\btheta^{(k)}\right\|_{1}\right) +\lambda_{0}\left(\left\|\btheta_{S}^{(0)}\right\|_{1}-\left\|\hat{\bDelta}_{S}^{(0)}\right\|_{2}+\left\|\hat{\bDelta}_{S^c}^{(0)}\right\|_{1}-\left\|\btheta_{S^c}^{(0)}\right\|_{1}-\left\|\btheta_{S}^{(0)}\right\|_{1}-\left\|\btheta_{S^c}^{(0)}\right\|_{1}\right)\\
\geq & \hat{\bDelta}^\top \hat{\bSig} \hat{\bDelta}+\frac{\lambda_{0}}{2}\left(\left\|\hat{\bDelta}_{S^c}^{(0)}\right\|_{1}-3\left\|\hat{\bDelta}_{S}^{(0)}\right\|_{1}\right) +\sum_{k=1}^{K} \frac{\lambda_{k}}{2}\left\|\hat{\bDelta}^{(k)}\right\|_{1}-2 \sum_{k=1}^{K}\lk h_{k}.
\end{aligned}
$$
with probability larger than $1-c_{1} \exp \left(-c_{2} \log p \right)$.

By the definition of $\wev$, we have $\hat{\bDelta}^{(0)}=\wev-\sum_{k=1}^K \frac{n_{S} }{N} \hat{\bDelta}^{(k)}$. Therefore, applying triangle inequality yields
\begin{align}
\label{before_trans}
F(\hat{\bDelta}) &\geq  \hat{\bDelta}^\top \hat{\bSig} \hat{\bDelta}+\frac{1}{2}\lambda_{0}\left\|\wev_{S^{c}}\right\|_{1} - \frac{1}{2}\lambda_{0}\sum^{K}_{k=1} \frac{n_{S}}{N}\left\|\hat{\bDelta}_{S^c}^{(k)}\right\|_{1}-\frac{3}{2}\lambda_0 \left\|\wev_{S}\right\|_{1}- \frac{3}{2}\lambda_0 \sum_{k=1}^{K} \frac{n_{S}}{N}\left\|\hat{\bDelta}^{(k)}_{S}\right\|_{1} \nonumber\\
&\quad + \sum_{k=1}^{K} \frac{\lambda_{k}}{2}\left\|\hat{\bDelta}^{(k)}\right\|_{1}-2 \sum_{k=1}^{K}\lk h_{k}.
\end{align}

Recall that we select $\lambda_{0}, \dots,\lk$ such that $\frac{\lambda_{k}}{2} \geq \frac{3}{2} \frac{n_{S}}{N}\lambda_{0}$, so we have
\begin{align}
\label{lambda_ineq}
\sum_{k=1}^{K} \frac{\lambda_{k}}{2}\left\|\hat{\bDelta}^{(k)}\right\|_{1}- \frac{3}{2}\lambda_0 \sum_{k=1}^{K} \frac{n_{S}}{N}\left\|\hat{\bDelta}^{(k)}_{S}\right\|_{1}- \frac{1}{2}\lambda_{0}\sum^{K}_{k=1} \frac{n_{S}}{N}\left\|\hat{\bDelta}_{S^c}^{(k)}\right\|_{1} \geq 0.
\end{align}

Notice that $\hat{\btheta}$ is the solution to the problem (\ref{transobj}). We then have $\hat{\bDelta}=\hat{\btheta} - \btheta^{*} = \underset{\bDelta}{\operatorname{argmin}}  F(\bDelta)$. Since $F(\boldsymbol{0})=0$, it follows that $F(\hat{\bDelta}) \leq 0$. This summing with (\ref{before_trans}) and (\ref{lambda_ineq}) leads to
\begin{align}
\label{unifiedproperty1}
0 &\geq \hat{\bDelta}^\top \hat{\bSig} \hat{\bDelta}-\frac{3}{2}\lambda_0 \left\|\wev_{S}\right\|_{1}+\frac{1}{2}\lambda_{0}\left\|\wev_{S^{c}}\right\|_{1} -2\sum_{k=1}^{K}\lk h_{k} \\
\label{unifiedproperty2}
&=\hat{\bDelta}^\top \hat{\bSig} \hat{\bDelta}-\frac{3}{2}\lambda_0 \left\|\wev\right\|_{1}+2\lambda_{0}\left\|\wev_{S^{c}}\right\|_{1} -2\sum_{k=1}^{K}\lk h_{k}.
\end{align}

We now establish the upper bound for error measured in $\ell_2$ norm, $\|\wev\|_2$. An application of Lemma \ref{Thm1Lemma2} on (\ref{unifiedproperty1}) yields that with probability larger than $1 - c_{1}\exp(c_2 n_{T}) - c_{3} \exp(c_4 \log p)$,
$$
0 \geq (1 - u_{n})\alpha_{\min} \left\| \wev\right\|_2^2-\frac{3}{2}\lambda_0 \sqrt{s}\left\|\wev\right\|_{2}-(2+v_{n}) \sum_{k=1}^{K}\lk h_{k}
$$
where we use the fact that  $ \left\|\wev_{S^c}\right\|_{1} \geq 0$.
If $u_{n}=o(1)$, we can show that for a sufficiently large $n_{S}$,
\begin{align}
\label{l2ResultInProof}
\left\|\wev\right\|_{2} &\leq \frac{\frac{3}{2} \lambda_0 \sqrt{s}+\sqrt{\frac{9}{4} \lambda_0^2 s+4\left(1-u_{n}\right)\left(2+v_{n}\right)\alpha_{\min}\sum_{k=1}^K \lambda_k h_k}}{2\left(1-u_{n}\right) \alpha_{\min}} \nonumber\\
&\lesssim \sqrt{\frac{s \log p}{N}}+ \sqrt{ (1+v_{n})\sum_{k=1}^K \frac{n_{S}}{N} \sqrt{\frac{\log p}{n_{S}}} h_{k}}
\end{align}
by plugging in the choice of $\lz$ and $\lk$s, which is the desired result.

It remains to show the order of $v_{n}$ and prove that $u_{n}=o(1)$ under the conditions of Theorem \ref{stepone}. By the assumptions in Theorem 1 and the choice of $\lambda_{0}, \dots,\lambda_{K}$, we have
$$
\begin{aligned}
u_{n}=\frac{ 256\beta_{\max }\lz^2}{\alpha_{\min}\lambda^2_{k}\wedge(\lambda^2_{0}/(K+1)) }\frac{s \log p}{N}   \lesssim \frac{s \log p}{ n_{S}}=o(1),
\end{aligned}
$$
and
$$
v_{n} = \frac{256 \beta_{\max}}{\lambda^2_{k}\wedge(\lambda^2_{0}/(K+1)) } \frac{\log p}{N} \left(\sum_{k=1}^{K}\lk h_{k}\right)\lesssim \sqrt{\frac{K^2\log p}{n_{S}}}\bar{h}.
$$
The proof is then finished.

\subsection{Proof of Theorem 2}
Similar to the arguments in the proof of Theorem \ref{stepone}, we
first define $\tilL (\bd)=\frac{1}{2n_{T}}\left\|\by^{(0)}-\bX^{(0)} \hat{\bw}-\bXz \bd\right\|_2^2$ and $\tilde{F}(\bDelta)=\tilL\left(\bd^{*}+\bDelta\right)-\tilL\left(\bd^{*}\right)+\tilde{\lambda}\left\|\bd^{*}+\bDelta\right\|_1-\tilde{\lambda}\left\|\bd^{*}\right\|_1$, where $\bd^{*} =  \bz - \bw$ is the contrast between the target parameter and the averaged parameter. 
Denoting $\hat{\bDelta}^{\bd}=\hat{\bd}-\bd^{*}$. Recall that $\wev = \hat{\bw} - \bw$, by Hölder inequality and triangle inequality,
$$
\begin{aligned}
\left\langle\nabla \tilL\left(\bd^{*}\right), \hat{\bDelta}^{\bd}\right\rangle & =\frac{1}{n_{T}}\left\langle \left(\bXz\right)^{\top}\left[\by^{(0)}-\bXz \bb^{(0)}-\bXz\left(\hat{\bw}+\bd^{*}-\bb^{(0)}\right)\right], \hat{\bDelta}^{\bd}\right\rangle \\
& =\frac{1}{n_{T}}\left\langle \left(\bXz\right)^{\top}\left[\bez-\bXz(\hat{\bw}-
\bw)\right], \hat{\bDelta}^{\bd}\right\rangle \\
& \leq \frac{1}{n_{T}}\left\|\left(\bXz\right)^{\top}\bez\right\|_{\infty}\|\hat{\bDelta}^{\bd} \|_1+\frac{1}{2} \left(\hat{\bDelta}^{\bd}\right)^{\top} \hat{\bSig}^{(0)} \hat{\bDelta}^{\bd}+\frac{1}{2} (\wev)^{\top} \hat{\bSig}^{(0)} \wev .
\end{aligned}
$$

By Lemma \ref{concentration}, if $n_{T} \gtrsim \log p$, we can choose $\tilde{\lambda}=c \sqrt{\frac{\log p}{n_{T}}}$ for some constant $c$ so that $\frac{1}{n_{T}}\left\|\left(\bXz \right)^{\top}\bez\right\|_{\infty} \leq \frac{\tilde{\lambda}}{2}$ with probability larger than $1-c_{1}\exp(c_{2} \log p)$. Therefore, it holds that
$$
\begin{aligned}
\tilL\left(\hat{\bDelta}^{\bd}+\bd^{*}\right)-\tilL(\widehat{\bDelta}^{\delta}) & =\left\langle\nabla \tilL\left(\bd^{*}\right), \hat{\bDelta}^{\bd}\right\rangle+\left(\hat{\bDelta}^{\bd}\right)^{\top} \hat{\bSig}^{(0)} \hat{\bDelta}^{\bd} \\
& \geq-\frac{\tilde{\lambda}}{2}\|\hat{\bDelta}^{\bd}\|_{1}+\frac{1}{2} \left(\hat{\bDelta}^{\bd}\right)^{\top} \hat{\bSig}^{(0)} \hat{\bDelta}^{\bd}-\frac{1}{2}(\wev)^{\top} \hat{\bSig}^{(0)} \wev
\end{aligned}
$$

So by the optimality condition,
\begin{align}
\label{discusscase}
0 & \geq \tilde{F}(\hat{\bDelta}^{\bd}) \nonumber\\
& \geq \tilL\left(\hat{\bDelta}^{\bd}+\bd^{*}\right)-\tilL(\hat{\bDelta}^{\bd})+\tilde{\lambda}\left\|\hat{\bDelta}^{\bd}+\bd^{*}\right\|_1-\tilde{\lambda}\left\|\bd^{*}\right\|_1 \nonumber\\
& \geq \tilL\left(\hat{\bDelta}^{\bd}+\bd^{*}\right)-\tilL(\hat{\bDelta}^{\bd})+\tilde{\lambda}\|\hat{\bDelta}^{\bd}\|_1-2 \tilde{\lambda}\left\|\bd^{*}\right\|_1 \nonumber\\
& \geq \frac{\tilde{\lambda}}{2}\|\hat{\bDelta}^{\bd}\|_1+\frac{1}{2} \left(\hat{\bDelta}^{\bd}\right)^{\top} \hat{\bSig}^{(0)} \hat{\bDelta}^{\bd}-\frac{1}{2}(\wev)^{\top} \hat{\bSig}^{(0)} \wev-2 \tilde{\lambda}\left\|\bd^{*}\right\|_1
\end{align}

Since the result involves $\wev$, we first establish the following auxiliary lemma:

\begin{lemma}
\label{steponel1bound}
    Under Assumption \ref{A1} and \ref{A2}, $n_{S} \gtrsim \log p$, $n_{S} > n_{T}$, if we choose $\lz \gtrsim \sqrt{\frac{\log p}{N}}$, $\lk = a_{k}\lz \gtrsim \sqrt{\frac{n_{S}}{N}}\sqrt{\frac{\log p}{N}}$ such that 
    $$\lk \geq 8 \lz \frac{n_{S}}{N}, \text{ }u_{n}=\frac{ 256\beta_{\max }\lz^2}{\alpha_{\min}\lambda^2_{k}\wedge(\lambda^2_{0}/(K+1)) }\frac{s \log p}{N}=o(1) \text{ and } v_{n} =\frac{256 \beta_{\max}}{\lambda^2_{k}\wedge(\lambda^2_{0}/(K+1)) } \frac{\log p}{N} \left(\sum_{k=1}^{K}\lk h_{k}\right)= O(1),$$
    then we have
    \begin{align*}
    \|\wev\|_{2}& \lesssim \sqrt{s}\lz + \sqrt{\sum^{K}_{k=1} \lk h_{k}} \\
    \|\wev\|_{1}& \lesssim s \lambda_{0} + \sqrt{s} \sqrt{\sum^{K}_{k=1} \lk h_{k}} + \frac{\sum^{K}_{k=1} \lk h_{k}}{\lz}  
    \end{align*}
    with probability larger than $1 - c_{1}\exp(c_2 n_{T}) - c_{3} \exp(c_4 \log p)$.
\end{lemma}

Notice that according to the theorem statement, the choice of $\lz$ and $\lk$s depends on the event $A$. It can be verified that either selection fulfills the conditions outlined in Lemma \ref{steponel1bound}. We now discuss by cases and apply Lemma \ref{steponel1bound} to prove the result.

\paragraph{Case 1:}
We start with the case when the event $A$ holds. That is, we have 
\begin{align}
\label{Aevent}
s\log p / n_{S} \geq \bar{h} \sqrt{\log p /n_{T}} 
\end{align}
Under this condition, we choose $$
 \lambda_{0} = c_{1} \sqrt{\frac{\log p}{N}} \text{, and } a_{k} = 8\sqrt{\frac{n_{S}}{N}}.
$$ 
Applying Lemma \ref{steponel1bound} yields
\begin{align}
&\|\wev\|_{2}\lesssim \sqrt{\frac{s \log p}{N}}+ \sqrt{  \sqrt{\frac{\log p}{n_{S}}} \bar{h}} \label{l2normcase1} \\
&\|\wev\|_{1}\lesssim s \sqrt{\frac{\log p}{N}} + \sqrt{  \sqrt{\frac{\log p}{n_{S}}} s\bar{h}} + \sqrt{\frac{N}{n_{S}}}\bar{h} \label{l1normcase1}
\end{align}
with probability larger than $1 - c_{1}\exp(c_2 n_{T}) - c_{3} \exp(c_4 \log p)$.

(i). If $\frac{1}{2}(\wev)^{\top} \hat{\bSig}^{(0)} \wev \geq 2 \tilde{\lambda}\left\|\bd^{*}\right\|_{1}$, according to (\ref{discusscase}), we have
\begin{align}
\label{condition1}
0 \geq \frac{\tilde{\lambda}}{2}\|\hat{\bDelta}^{\bd}\|_{1}+\frac{1}{2} \left(\hat{\bDelta}^{\bd}\right)^{\top} \hat{\bSig}^{(0)} \hat{\bDelta}^{\bd}-(\wev)^{\top} \hat{\bSig}^{(0)} \wev.   
\end{align}

By Lemma \ref{RSC} and the condition $n_{T} \gtrsim \log p$, we have $(\wev)^{\top} \hat{\bSig}^{(0)} \wev \leq \gamma_0\left\|\wev\right\|_2^2 + \tau_{0} \frac{\log p}{n_{T}} \left\|\wev\right\|_1^2$ for some constants $\gamma_{0}$ and $\tau_{0}$, this together with (\ref{condition1}) indicates
\begin{align}
\label{steptwol1bound}
\frac{\tilde{\lambda}}{2}\|\hat{\bDelta}^{\bd}\|_1 &\leq (\wev)^{\top} \hat{\bSig}^{(0)} \wev
\leq \gamma_0\left\|\wev\right\|_2^2 + \tau_{0} \frac{\log p}{n_{T}} \left\|\wev\right\|_1^2 \nonumber\\
&\lesssim \frac{s \log p}{N}+(1+\frac{s\log p}{n_{T}}) \sqrt{\frac{\log p}{n_{S}}} \bar{h} + \frac{(s \log p)^2}{n_{T}N} + \frac{N}{n_{S}}\bar{h}^2 \frac{\log p}{n_{T}}.
\end{align}
where the last inequality is based on the results in   (\ref{l2normcase1}) and (\ref{l1normcase1}). In Theorem \ref{steptwo} we assume $\frac{s \log p}{n_{T}} = O(1)$ and $\bar{h}\sqrt{\log p/n_{T}} = o(1)$. According to (\ref{Aevent}), we have $(N/n_{S})\bar{h}\sqrt{\log p / n_{T}} \le (K+1)s\log p / n_{S} = O(1)$. Applying these results to (\ref{steptwol1bound}) yields  $\frac{\tilde{\lambda}}{2}\|\hat{\bDelta}^{\bd}\|_1 = o_{p}(1)$.

On the other hand, if we apply Lemma \ref{RSC} to (\ref{condition1}), we then have 
$$
0 \geq \left(1 - \frac{\beta_{0}}{\tilde{\lambda}}\frac{{\log p}}{n_{T}}\|\hat{\bDelta}^{\bd}\|_{1}\right)\frac{\tilde{\lambda}}{2}\|\hat{\bDelta}^{\bd}\|_{1}+\frac{1}{2} \alpha_{0}\|\hat{\bDelta}^{\bd}\|_{2}^{2}-\gamma_{0}\left\|\wev\right\|_{2}^{2}- \tau_{0} \frac{\log p}{n_{T}} \left\|\wev\right\|_1^2.
$$

Notice that we choose $\tilde{\lambda}=c\sqrt{\frac{\operatorname{\log p}}{n_{T}}}$ for some universal constant $c$. Therefore, we may choose $c > 
\sqrt{2\beta_{0}}$ so we have $ \beta_{0}\log p / n_{T}<\tilde{\lambda}^2/2$. This together with (\ref{steptwol1bound}) and the fact that $\frac{\tilde{\lambda}}{2}\|\hat{\bDelta}^{\bd}\|_1 = o_{p}(1)$ leads to
$$
\frac{1}{2} \alpha_{0}\|\hat{\bDelta}^{\bd}\|_{2}^{2} \leq \gamma_{0}\left\|\wev\right\|_{2}^{2}+ \tau_{0} \frac{\log p}{n_{T}} \left\|\wev\right\|_1^2.
$$
Based on similar arguments to those in (\ref{steptwol1bound}), we have
$$
\|\hat{\bDelta}^{\bd}\|_{2} \lesssim \sqrt{\frac{s \log p}{N}}+\sqrt{ \sqrt{\frac{\log p}{n_{S}} }\bar{h}} + \sqrt{\frac{N}{n_{S}}}\sqrt{\frac{\log p}{n_{T}}}\bar{h}.
$$
with probability larger than $1 - c_{1}\exp(c_2 n_{T}) - c_{3} \exp(c_4 \log p)$.

(ii). If $\frac{1}{2}(\wev)^{\top} \hat{\bSig}^{(0)} \wev \leq 2 \tilde{\lambda}\left\|\bd^{*}\right\|_{1}$, we have
$$
0 \geq \frac{\tilde{\lambda}}{2}\|\hat{\bDelta}^{\bd}\|_1+\frac{1}{2} \left(\hat{\bDelta}^{\bd}\right)^{\top} \hat{\bSig}^{(0)} \hat{\bDelta}^{\bd}-4 \tilde{\lambda}\left\|\bd^{*}\right\|_1
$$
which implies that $\|\hat{\bDelta}^{\bd}\|_1 \leq 8 \|\bd^{*}\|_1 \leq 8\bar{h}$.

By applying Lemma \ref{RSC} again, we have
$$
\begin{aligned}
& 0 \geq \frac{\tilde{\lambda}}{2}\|\hat{\bDelta}^{\bd}\|_{1}+\frac{1}{2} \alpha_{0}\|\hat{\bDelta}^{\bd}\|_{2}^{2}-\frac{1}{2} \beta_{0} \frac{\log p}{n_{T}}\|\hat{\bDelta}^{\bd}\|_{1}^{2}-4 \tilde{\lambda}\left\|\bd^{*}\right\|_{1} \\
& \geq \frac{1}{2} \alpha_{0}\|\hat{\bDelta}^{\bd}\|_{2}^{2}-32 \beta_{0} \frac{\log p}{n_{T}}\left\|\bd^{*}\right\|_{1}^{2}-4 \tilde{\lambda}\left\|\bd^{*}\right\|_{1} .
\end{aligned}
$$
with probability larger than $1 - c_{1}\exp(c_2 n_{T})$.

So in this case, we have
$$
\begin{aligned}
\|\hat{\bDelta}^{\bd}\|_{2} \leq \sqrt{\frac{64 \beta_{0}}{\alpha_{0}} \frac{\log p}{n_{T}} \bar{h}^{2}+8 \frac{\tilde{\lambda}}{\alpha_{0}} \bar{h}} \lesssim \sqrt{\frac{\log p}{n_{T}}} \bar{h}+\sqrt{\sqrt{\frac{\log p}{n_{T}} } \bar{h}}
\end{aligned}
$$
and
$$
\|\hat{\bDelta}^{\bd}\|_2 \leq \|\hat{\bDelta}^{\bd}\|_1 \leq 8\bar{h}
$$

Under the assumption that $\bar{h} \sqrt{\frac{\log p}{n_{T}}}=o(1)$, we have
$
\|\hat{\bDelta}^{\bd}\|_{2}  \lesssim \sqrt{\sqrt{\frac{\log p}{n_{T}} } \bar{h}} \wedge \bar{h} .
$

Therefore, by combining the results from the two cases discussed above, we have
$$
\left\|\hat{\bw}+\hat{\bd}-\bb^{(0)}\right\|_{2} \leq \left\|\hat{\bDelta}^{\bd}\right\|_{2}+\left\|\wev\right\|_{2} \lesssim \sqrt{\frac{s \log p}{N}}+\sqrt{ \sqrt{\frac{\log p }{n_{S}}} \bar{h}} + \sqrt{K+1}\sqrt{\frac{\log p}{n_{T}}}\bar{h}+\sqrt{\sqrt{\frac{\log p}{n_{T}} }\bar{h}} \wedge \bar{h}
$$
with probability larger than $1 - c_{1}\exp(c_2 n_{T}) - c_{3} \exp(c_4 \log p)$.

Since $A$ holds, together with the condition $K^2 s \log p / n_{S} = O(1)$, we have 
$$
\sqrt{K+1}\sqrt{\frac{\log p}{n_{T}}}\bar{h} \leq \sqrt{K+1} \frac{s \log p}{n_{S}} \le  \sqrt{\frac{(K+1)^2s \log p}{n_{S}}}\sqrt{\frac{s \log p}{N}} \lesssim \sqrt{\frac{s \log p}{N}}
$$
which implies
$$
\left\|\hat{\bw}+\hat{\bd}-\bb^{(0)}\right\|_{2} \leq \left\|\hat{\bDelta}^{\bd}\right\|_{2}+\left\|\wev\right\|_{2} \lesssim \sqrt{\frac{s \log p}{N}}+\sqrt{ \sqrt{\frac{\log p }{n_{S}}} \bar{h}} +\sqrt{\sqrt{\frac{\log p}{n_{T}} }\bar{h}} \wedge \bar{h}.
$$

\paragraph{Case 2:}
Next we discuss the case when the event $A^c$ holds, i.e.,
\begin{align*}
s\log p / n_{S} \leq \bar{h} \sqrt{\log p /n_{T}} 
\end{align*}
In this case, we choose 
$$
 \lambda_{0} = c_{0} \sqrt{\frac{\log p}{n_{S}}}  \text{, and } a_{k} =  \frac{8n_{S}}{N}
$$ 
Applying Lemma \ref{steponel1bound} again, we have
\begin{align}
&\|\wev\|_{2}\lesssim \sqrt{\frac{s \log p}{n_{S}}}+ \sqrt{  \sqrt{\frac{\log p}{n_{S}}} \bar{h}} \label{l2normcase2} \\
&\|\wev\|_{1}\lesssim s \sqrt{\frac{\log p}{n_{S}}} + \sqrt{  \sqrt{\frac{\log p}{n_{S}}} s\bar{h}} +\bar{h} \label{l1normcase2}
\end{align}
with probability larger than $1 - c_{1}\exp(c_2 n_{T}) - c_{3} \exp(c_4 \log p)$.

Plugging the new bound (\ref{l1normcase2})
 and (\ref{l2normcase2}) into the arguments in Case 1 leads to 
$$
\left\|\hat{\bw}+\hat{\bd}-\bb^{(0)}\right\|_{2}  \lesssim \sqrt{\frac{s \log p}{n_{S}}}+\sqrt{ \sqrt{\frac{\log p }{n_{S}}} \bar{h}} + \sqrt{\frac{\log p}{n_{T}}}\bar{h}+\sqrt{\sqrt{\frac{\log p}{n_{T}} }\bar{h}} \wedge \bar{h}.
$$

Recall that in Theorem \ref{steptwo} we assume $\sqrt{\log p/n_{T}}\bar{h} = o(1)$ and $\log p / n_{T} = O(1)$. Therefore, in the above bound, the third term has a smaller order comparing to the fourth term. Hence, we have 

$$
\left\|\hat{\bw}+\hat{\bd}-\bb^{(0)}\right\|_{2}  \lesssim \sqrt{\frac{s \log p}{n_{S}}}+\sqrt{ \sqrt{\frac{\log p }{n_{S}}} \bar{h}}+\sqrt{\sqrt{\frac{\log p}{n_{T}} }\bar{h}} \wedge \bar{h}.
$$

As we assume $A^c$ holds in this case, we have $s \log p / n_{S} \le \bar{h}\sqrt{\log p / n_{T}}$, which further implies
$$
\left\|\hat{\bw}+\hat{\bd}-\bb^{(0)}\right\|_{2}  \lesssim \sqrt{\sqrt{\frac{\log p}{n_{T}} }\bar{h}}.
$$

Combining the results from the two cases discussed above, we have 
$$
\left\|\hat{\bw}+\hat{\bd}-\bb^{(0)}\right\|_{2}  \lesssim \sqrt{\frac{s \log p}{N}}+\sqrt{\sqrt{\frac{\log p}{n_{T}} }\bar{h}}.
$$
with probability larger than $1 - c_{1}\exp(c_2 n_{T}) - c_{3} \exp(c_4 \log p)$. The proof is then completed.

\subsection{Proof of Theorem \ref{ComTransFusion}}

We first show a lemma discussing the bias and variance components in (\ref{biasvarcomp}), whose proof is based on the results in ~\cite{javanmard2014confidence} but taking into account that $\bk$ is not exactly $s$-sparse (due to the contrast term $\bdk$).

\begin{lemma}
\label{DebiasBound}
Under Assumption \ref{A1} and \ref{A2} and $ \frac{s \log p}{n_S}=o(1)$, if we construct $\{\hat{\bb}^{(k)}_{LASSO}\}_{k = 1, \dots, K}$ through (\ref{Comobj}) and $\{\hThetak\}_{k=1,\dots,K}$ using (\ref{JMalgorithm}), with parameters $\tilde{\lambda}_k = \mu_{k} =c_{0}\sqrt{\frac{\log p}{n_S}}$ for some universal constant $c_{0}$, then we have that for $k = 1, \dots, K$,
\begin{align}
\label{VarPart}
\frac{1}{n_{S}}\left\|\hThetak\left(\bX^{(k)}\right)^{\top} \boldsymbol{\epsilon}^{(k)}\right\|_{\infty} \lesssim \sqrt{\frac{\log p}{n_{S}}} 
\end{align}
and 
\begin{align}
\label{biastermbound}
\left\|\boldsymbol{b}^{(k)}\right\|_{\infty} := \left\|\left(\hThetak \hat{\boldsymbol{\Sigma}}^{(k)}-\boldsymbol{I}\right)\left(\hat{\bb}^{(k)}_{\LASSO}-\bk\right)\right\|_{\infty} \lesssim \frac{s \log p}{n_{S}}+h_{k} \sqrt{\frac{\log p}{n_{S}}},
\end{align}
with probability larger than $1- c_{1}\exp(-c_{2}\log p)$.
\end{lemma}

Now we proceed to the proof of the theorem. We first show that by reparametrization, problem (\ref{Comobj}) is essentially a special case of problem (\ref{obj}). Then we apply techniques similar to those used in Theorem \ref{stepone} to prove the results.

Following the arguments in (\ref{ULMForm}), we may reformulate problem (\ref{Comobj}) into a generalized LASSO problem: 
\begin{align}
\label{ComForm}
    \tilde{\btheta} = \underset{\btheta}{\operatorname{argmin}} \left\{ \tilde{L}(\btheta) + \lambda_{0}\mR(\btheta) \right\}
     := \underset{\btheta}{\operatorname{argmin}}\left\{\frac{1}{2N}\left\|\tilde{\by}-\tilde{\bX} \btheta\right\|_{2}^{2}+\lambda_{0}\mR(\btheta)  \right\}
\end{align}
where
\begin{align}
\label{ULMForm}
\tilde{\by}=\left(\begin{array}{c}
\sqrt{n_S}\tilde{\bb}^{(1)} \\
\sqrt{n_S}\tilde{\bb}^{(2)}  \\
\vdots \\
\sqrt{n_{S}}\tilde{\bb}^{(K)}  \\
\by^{(0)}
\end{array}\right), \quad  \tilde{\bX}=\left(\begin{array}{ccccc}
\sqrt{n_S}\boldsymbol{I}_{p} & 0 & \cdots & 0 & \sqrt{n_S}\boldsymbol{I}_{p} \\
0 & \sqrt{n_S}\boldsymbol{I}_{p} & \cdots & 0 & \sqrt{n_S}\boldsymbol{I}_{p} \\
\vdots & \vdots & \vdots & \vdots & \vdots \\
0 & 0 & \cdots & \sqrt{n_{S}}\boldsymbol{I}_{p} & \sqrt{n_{S}}\boldsymbol{I}_{p} \\
0 & 0 & \cdots & 0 & \bXz
\end{array}\right).
\end{align}
Similarly, we can define the random noise $\tilde{\boldsymbol{\epsilon}} = \tilde{\by} - \tilde{\bX}\btheta^{*}$. The k-th block of $\tilde{\boldsymbol{\epsilon}}$ is given by $\sqrt{n_{S}}\left(\hThetak\left(\bX^{(k)}\right)^{\top} \bek/n_{S} + \boldsymbol{b}^{(k)}\right)$ for $k=1,\dots, K$, whereas the last block is $\bez$, the observation noise for the target model.

Following the aforementioned reformulation, we can employ an approach similar to that used in Lemma \ref{concentration} to establish the result about $\left\langle\nabla \tilde{\mL}\left(\btheta\right), \bDelta\right\rangle$. Define $\delta_k = \frac{s \log p}{N} + \frac{n_{S}}{N} \sqrt{\frac{\log p}{n_{S}}}h_{k}$ and $\delta_0 =\frac{Ks \log p}{N} +\sqrt{\frac{\log p}{n_{S}}}\bar{h}$, the result is stated as follows:
\begin{lemma}
\label{ComConcentrastion}
    Under assumptions \ref{A1} and \ref{A2}, if $n_S \gtrsim \log p$, $\lk=c_{k} \left(\sqrt{\frac{n_{S}}{N} \frac{\log p}{N}}+\delta_{k} \right)$ and $\lambda_{0}=c_{0} \left(\sqrt{\frac{\log p}{N}} + \delta_{0} \right) $ for some appropriate constants $c_{0}, \dots, c_{K}$, then we have for any $\bDelta = \left(\left(\bDelta^{(1)}\right)^{\top}, \dots, \left(\bDelta^{(K)}\right)^{\top}, \left(\bDelta^{(0)}\right)^{\top} \right)^{\top}\in \mathbb{R}^{(K+1) p}$,
$$
\left|\left\langle\nabla \tilde{L}\left(\btheta\right), \bDelta\right\rangle\right|
\leq \sum_{k=1}^{K} \frac{\lambda_{k}}{2}\left\|\bDelta^{(k)}\right\|_{1}+\frac{\lambda_{0}}{2}\left\|\bDelta^{(0)}\right\|_{1} .
$$
with probability larger than $1-c_{1} \exp \left(-c_{2} \log p \right)$.
\end{lemma}

Notice that the only difference between Lemma \ref{ComConcentrastion} and Lemma \ref{concentration} is the choice of $\{\lambda_{k}\}_{k=0 \le k \le K}$. With this new choice of parameters, we can verify that if further $n_{S} \gg Ks\log p$, $n_{S} \gtrsim K^2 \log p$, and $h_{k} \asymp \bar{h}$ for any $1 \le k \le K$, the conditions of Lemma \ref{steponel1bound} hold. Therefore, we can apply Lemma \ref{steponel1bound} and obtain
\begin{align}
\label{D-TransFusionproofbound}
  \|\hat{\bw}_{C} - \bw\|_{2}  \lesssim \sqrt{s}\lambda_{0} +\sqrt{\sum^{K}_{k=1}\lambda_{k}h_{k}}  \lesssim \sqrt{s}\left(\sqrt{\frac{ \log p}{N}} + \delta_{0}\right)+\sqrt{\sum_{k=1}^K \left(\frac{n_{S}}{N} \sqrt{\frac{\log p}{n_{S}}} + \delta_k \right) h_{k}}.
\end{align}
with probability larger than $1 - c_{1}\exp(c_2 n_{T}) - c_{3} \exp(c_4 \log p)$. This together with the fact that $\|\bw - \bz\|_{2}^2 \le \epsilon_{D}^2$ implies the bound (\ref{D-TransFusionbound1}). 

Furthermore, if we assume $h_{k} \asymp \bar{h} = O(1)$, then we have
\begin{align}
\label{negligibleproof1}
s \delta_0^2 = \frac{K^2s^3 \log^2 p}{N^2} + \frac{s\log p}{n_{S}}\bar{h}^2\lesssim \frac{s \log p}{N} + \sqrt{\frac{\log p}{n_{S}}}\bar{h}
\end{align}
and
\begin{align}
\label{negligibleproof2}
    \sum^{K}_{k=1}\delta_k h_k = \frac{s \log p}{n_{S}}\bar{h} + \sqrt{\frac{\log p}{n_{S}}} \sum^{K}_{k=1}\frac{n_{S}}{N}h_k^2 \lesssim \sqrt{\frac{\log p}{n_{S}}}\bar{h}
\end{align}
based on the condition that $n_S \gg Ks^2\log p$. Therefore, we have the bound (\ref{D-TransFusionbound2}).

\subsection{Proof of Theorem \ref{ComStep2}}
We use a similar line of arguments as the proof of Theorem \ref{steptwo}. Recall that we choose
\begin{align}
\lk&=c_{0}(8 \vee \frac{\bar{h}}{h_{k}})  \left(\sqrt{\frac{n_{S
    }}{N} \frac{\log p}{N}}+\delta_{k} \right), \delta_k = \frac{s \log p}{N} + \frac{n_{S}}{N} \sqrt{\frac{\log p}{n_{S}}}h_{k}, \\
    \lz&=c_{0} \left(\sqrt{\frac{\log p}{N}} \mathbb{1}_{A} + \sqrt{\frac{\log p}{n_{S}}} \mathbb{1}_{A^c} + \delta_{0} \right), 
\delta_0 =\frac{Ks \log p}{N} +\sum^{K}_{k=1} \frac{n_{S}}{N}\sqrt{\frac{\log p}{n_{S}}}h_{k}. \label{lambda_choice}
\end{align}
We can verify that if $n_S \gg Ks\log p$, $n_{S} \gtrsim K^2 \log p$ and $h_{k} \asymp \bar{h}$ for any $1 \le k \le K$, this choice of parameters satisfies the conditions in Lemma \ref{steponel1bound}. Therefore we can apply Lemma \ref{steponel1bound} and obtain
\begin{align}
    \|\wev\|_{2}& \lesssim \sqrt{s}\lz + \sqrt{\sum^{K}_{k=1} \lk h_{k}} \lesssim \sqrt{{\frac{s \log p}{N}}}\mathbb{1}_{A}+\sqrt{{\frac{s \log p}{n_S}}}\mathbb{1}_{A^c}+ \sqrt{\sqrt{\frac{\log p}{n_S}}\bar{h}} + \sqrt{s}\delta_{0} + \sqrt{\sum^{K}_{k=1} \delta_k  h_{k}} \label{generalbound1}\\
    \|\wev\|_{1}& \lesssim s \lambda_{0} + \sqrt{s} \sqrt{\sum^{K}_{k=1} \lk h_{k}} + \frac{\sum^{K}_{k=1} \lk}{\lz} \bar{h}\nonumber \\ 
    &\lesssim \sqrt{\frac{s^2 \log p}{N}}\mathbb{1}_{A} + \sqrt{\frac{s^2 \log p}{n_{S}}}\mathbb{1}_{A^c} + \sqrt{\sqrt{\frac{\log p}{n_S}}s\bar{h}} + s\delta_0 + \sqrt{\sum^K_{k=1}s\delta_k h_k} + \sqrt{K}\bar{h}\mathbb{1}_{A} + \bar{h}\mathbb{1}_{A^c} 
    \label{generalbound2}
\end{align}
by plugging in the choice of $\lz$ and $\lk$s and using the fact that $\sum^K_{k=1}\delta_k = \delta_0$.

To prove the theorem, it suffices to show that in the bounds (\ref{generalbound1}) and (\ref{generalbound2}), the terms involving $\delta_{0}$ and $\delta_{k}$ are of orders that align with some other terms, then we can follow exactly the same proof in Theorem \ref{steptwo} to prove the result. To show this, we can use the results in (\ref{negligibleproof1}) and (\ref{negligibleproof2}), which gives us
\begin{align}
    \sqrt{s}\delta_0 + \sqrt{\sum^K_{k=1}\delta_k h_k} \lesssim \sqrt{\frac{s \log p}{N}} + \sqrt{\sqrt{\frac{\log p}{n_{S}}}\bar{h}}.
\end{align}
With this, the proof is completed.

\section{Proof of Technical Lemmas}
\label{lemmaproof}

\subsection{Proof of Lemma \ref{concentration}} By definition,
$-\nabla \mL\left(\btheta\right)=\frac{1}{N} \bX^{\top}\left(\by-\bX \btheta\right)=\left(\frac{1}{N} \left(\bX^{(1)}\right)^{\top} \boldsymbol{\epsilon}^{(1)}, \dots, \frac{1}{N} \left(\bX^{(K)}\right)^{\top} \boldsymbol{\epsilon}^{(K)}, \frac{1}{N} \sum_{k=0}^{K}  \left(\bX^{(k)}\right)^{\top} \bek\right)^{\top}$.

Therefore, by Hölder's inequality, we have
$$
\begin{aligned}
\left|\left\langle\nabla \mL\left(\btheta\right), \bDelta\right\rangle\right| &=\sum_{k=1}^{K} \left| \left\langle\frac{1}{N} \left(\bX^{(k)}\right)^{\top} \bek, \bDelta^{(k)}\right\rangle \right| +\left|\left\langle\frac{1}{N}\sum_{k=0}^{K}  \left(\bX^{(k)}\right)^{\top} \bek, \bDelta^{(0)}\right\rangle \right| \\
& \leq \sum_{k=1}^{K}\left\|\frac{1}{N} \left(\bX^{(k)}\right)^{\top} \bek\right\|_{\infty}\left\|\bDelta^{(k)}\right\|_{1}+\left\|\frac{1}{N}\sum_{k=0}^{K}  \left(\bX^{(k)}\right)^{\top} \bek\right\|_{\infty}\left\|\bDelta^{(0)}\right\|_{1} .
\end{aligned}
$$

Recall that we define $n_{k} = n_{S}$ for $1 \le k \le K$ and $n_{k} = n_{T}$ for $k=0$. Define
$$
\mathcal{A}^{(k)}=\left\{\max _{\substack{1 \leq j \leq p}}\left\{\frac{1}{n_{k}} \sum_{i=1}^{n_{k}}\left(x_{i j}^{(k)}\right)^{2}\right\} \leq 2 \max_{1 \le j \le p} E\left(x_{i j}^{(k)}\right)^{2}\right\}, $$
and
$$
\mathcal{A}=\left\{\max _{\substack{1 \leq j \leq p}}\left\{\frac{1}{N} \sum^{K}_{k=0}\sum_{i=1}^{n_{k}}\left(x_{i j}^{(k)}\right)^{2}\right\} \leq 2 \max_{1 \le k \le K, 1 \le j \le p} E\left(x_{i j}^{(k)}\right)^{2}\right\}. $$
Since  $\bX^{(k)}$ is sub-Gaussian with uniformly bounded second moment and $n_{S} \gtrsim \log p$, we have $P\left(\overline{\mathcal{A}^{(k)}}\right) \leq c_{1} \exp (-c_{2} n_{S})$ for $1 \le k \le K$ and $P\left(\overline{\mathcal{A}}\right) \leq c_{1} \exp (-c_{2} N)$ for some universal constants $c_{1}$ and $c_{2}$.

In addition, as $\bek \sim N\left(0, \sigma_{k}^{2}\boldsymbol{I}\right)$ for some finite $\sigma_{k}$, by Proposition 5.10 in \cite{vershynin2010introduction}, we can establish that with some universal constant $c_{3}$, for $1 \le k \le K$,
\begin{align*}
P\left(\max _{1 \leq j \leq p}\left|\frac{1}{n_{S}} \sum_{i=1}^{n_{S}} \epsilon_{i}^{(k)} x_{i j}^{(k)}\right| \geq t\right) 
&\leq P\left(\max _{1 \leq j \leq p}\left|\frac{1}{n_{S}} \sum_{i=1}^{n_{S}} \epsilon_{i}^{(k)} x_{i j}^{(k)}\right| \geq t  \ \middle| \  \mathcal{A}^{(k)}\right) + P(\overline{\mathcal{A}^{(k)}}) \\
&\leq p \cdot e \cdot \exp \left(-\frac{c_3 n_{S} t^{2}}{4 \sigma_{k}^2 \max_{1 \le j \le p} E\left(x_{i j}^{(k)}\right)^{2}}\right) + c_{1}\exp{(-c_{2}n_{S})}.
\end{align*}

Since by Assumption \ref{A1}, there exist a constant $c$ such that $\max_{1 \leq k \leq K}\Lambda_{\max}(\Sigk) \leq c$, so $\max_{1 \le j \le p} E\left(x_{i j}^{(k)}\right)^{2}$ is uniformly bounded above. Therefore for $1 \le k \le K$, by choosing $t = \sqrt{c_{4} \log p / n_{S}}$ for some constant $c_{4}$,  with probability larger than $1-c_{1} \exp \left(-c_{2} \log p\right)$, we have
$$
\left\|\frac{1}{N} \left(\bX^{(k)}\right)^{\top} \bek\right\|_{\infty} \lesssim \frac{n_{S}}{N} \sqrt{\frac{\log p}{n_{S}}} = \sqrt{\frac{n_{S}}{N}}\sqrt{\frac{\log p}{N}} .
$$

Similarly, we have
\begin{align*}
P\left(\max _{1 \leq j \leq p}\left|\frac{1}{N} \sum_{k=0}^{K} \sum_{i=1}^{n_{k}} \epsilon_{i}^{(k)} x_{i j}^{(k)}\right| \geq t \right) 
&\leq P\left(\max _{1 \leq j \leq p}\left|\frac{1}{N} \sum_{k=0}^{K} \sum_{i=1}^{n_{k}} \epsilon_{i}^{(k)} x_{i j}^{(k)}\right| \geq t  \ \middle| \ \mathcal{A}\right) + P(\overline{\mathcal{A}})  \\
&\leq p \cdot e \cdot \exp \left(-\frac{c_{4} N t^{2}}{4 \max _ {0 \le k \le K, 1 \le j \le p}\sigma_{k}^2E\left(x_{ij}^{(k)}\right)^{2}}\right) +c_{1} \exp \left(-c_{2} N\right)
\end{align*}

So we have with probability larger than $1-c_{1} \exp \left(-c_{2} \log p \right)$,
$$
\left\|\sum_{k=0}^{K} \frac{1}{N} \left(\bX^{(k)}\right)^{\top} \bek\right\|_{\infty}
\lesssim \sqrt{\frac{\log p}{N}}
$$

Therefore, by choosing $\lambda_{k}=a_{k} \lk = c_{0} \sqrt{\frac{n_{S}}{N}} \sqrt{\frac{\log p}{N}}$ and $\lambda_{0}=c_{0} \sqrt{\frac{\log p}{N}}$ for some sufficiently large constant $c_{0}$, we have the desired result.

\subsection{Proof of Lemma \ref{Thm1Lemma1}} Define $S$ is the support set of $\btheta^{(0)}=\bb^{(0)}$, and $S^{c}$ as its complement. Then we have $\left\|\btheta_{S}^{*}\right\|_{0}=s$ and $\left\|\btheta_{S^c}^{*}\right\|_{1} \leq \sum^{K}_{k=1}h_{k} \leq (K+1)\bar{h}$ as $n_S \ge n_{T}$. 

We define $F: \mathbb{R}^{(K+1)p} \rightarrow \mathbb{R}$ as
$$F(\bDelta)=\mL\left(\btheta^{*}+\bDelta\right)-\mL\left(\btheta^{*}\right)+\lz \mR \left(\btheta^{*}+\bDelta\right)- \lz \mR \left(\btheta^{*}\right),$$
and $\hat{\btheta}$ as the solution to the problem (\ref{transobj}). We then have $\hat{\bDelta}=\hat{\btheta} - \btheta^{*} = \underset{\bDelta}{\operatorname{argmin}}  F(\bDelta)$ and $F(0)=0$. Consequently, it follows that $F(\hat{\bDelta}) \leq 0$.

Since $\mL$ is a convex function, by Lemma \ref{concentration}, we can choose $\lambda_{k} = c_0 \sqrt{\frac{n_{S}}{N}}\sqrt{\frac{\log p}{N}}$ and $\lambda_{0} = c_0 \sqrt{\frac{\log p}{N}}$ so that 
\begin{align}
\label{Convexity}
\mL\left(\btheta^{*}+\hat{\bDelta}\right)-\mL\left(\btheta^{*}\right) \geq \left\langle\nabla \mL\left(\btheta^{*}\right), \hat{\bDelta}\right\rangle
\geq -\sum_{k=1}^{K} \frac{\lambda_{k}}{2}\left\|\hat{\bDelta}^{(k)}\right\|_{1}-\frac{\lambda_{0}}{2}\left\|\hat{\bDelta}^{(0)}\right\|_{1}
\end{align}
with probability larger than $1 - c_{1} \exp(c_2 \log p)$.

Since the $\ell_1$-norm function is decomposable and $\|\btheta_{S^c}^{(0)}\|_1 = 0$, by triangle inequality we have

\begin{align}
\label{RestrictedCone}
\lz \mR \left(\btheta^{*}+\bDelta\right)- \lz \mR \left(\btheta^{*}\right)
=& \sum_{k=1}^{K}\lk\left(\left\|\btheta^{(k)}+\hat{\bDelta}^{(k)}\right\|_{1}-\left\|\btheta^{(k)}\right\|_{1}\right)+\lambda_{0}\left(\left\|\btheta^{(0)}+\hat{\bDelta}^{(0)}\right\|_{1}-\left\|\btheta^{(0)}\right\|_{1}\right) \nonumber \\
\geq & \sum_{k=1}^{K}\lk\left(\left\|\hat{\bDelta}^{(k)}\right\|_{1}-2\left\|\btheta^{(k)}\right\|_{1}\right)+\lambda_{0}\left(\left\|\hat{\bDelta}_{S^c}^{(0)}\right\|_{1}-\left\|\hat{\bDelta}_{S}^{(0)}\right\|_{1}-2\left\|\btheta_{S^c}^{(0)}\right\|_{1}\right) \nonumber \\
\geq & \sum_{k=1}^{K}\lk\left\|\hat{\bDelta}^{(k)}\right\|_{1}-2 \sum_{k=1}^{K}\lk h_{k}+\lambda_{0}\left(\left\|\hat{\bDelta}_{S^c}^{(0)}\right\|_{1}-\left\|\hat{\bDelta}_{S}^{(0)}\right\|_{1}\right).
\end{align}

Combining (\ref{Convexity}) and (\ref{RestrictedCone}) yields
\begin{align}
\label{beforecone}
0 \geq F(\hat{\bDelta}) \geq \sum_{k=1}^{K} \frac{\lambda_{k}}{2}\left\|\hat{\bDelta}^{(k)}\right\|_{1}-2 \sum_{k=1}^{K}\lk h_{k}+\frac{\lambda_{0}}{2}\left(\left\|\hat{\bDelta}_{S^c}^{(0)}\right\|_{1}-3\left\|\hat{\bDelta}_{S}^{(0)}\right\|_{1}\right)
\end{align}

which leads to the following inequality:
$$
\sum_{k=0}^{K}\lk\left\|\hat{\bDelta}^{(k)}\right\|_{1}+\lambda_{0}\left\|\hat{\bDelta}^{(0)}\right\|_{1} \leq 4 \lambda_{0}\left\|\hat{\bDelta}_{S}^{(0)}\right\|_{1}+4 \sum_{k=1}^{K}\lk h_{k}.
$$

Recalling that we define $\wev=\sum_{k=1}^K \frac{n_S }{N} \hat{\bDelta}^{(k)}+\hat{\bDelta}^{(0)}$, it follows that
\begin{align}
\label{Cone}
\sum_{k=1}^K \lambda_k\left\|\hat{\bDelta}^{(k)}\right\|_1+\lambda_0\left\|\hat{\bDelta}^{(0)}\right\|_1 & \leq 4 \lambda_0\left\|\hat{\bDelta}_S^{(0)}\right\|_1+4 \sum_{k=1}^K \lambda_k h_{k} \nonumber \\
& \leq 4 \lambda_0\left\|\wev_S\right\|_1+4\sum_{k=1}^K \lambda_{0}\frac{n_S }{N} \left\|\hat{\bDelta}_S^{(k)}\right\|_{1}+4 \sum_{k=1}^K \lambda_k h_{k}
\end{align}

Since by the choice of parameters, we have 
$
\frac{\lambda_k}{2} \geq  4\lambda_0 \frac{n_S}{N}
$, so reorganizing (\ref{Cone}) yields
\begin{align}
\label{WeightedRSD}
\sum_{k=1}^K \lambda_k\left\|\hat{\bDelta}^{(k)}\right\|_1+2\lambda_0\left\|\hat{\bDelta}^{(0)}\right\|_1 \leq 8 \lambda_0\left\|\wev_S\right\|_1+8 \sum_{k=1}^K \lambda_k h_{k}
\end{align}
which is as desired.

\subsection{Proof of Lemma \ref{Thm1Lemma2}}
Applying Lemma \ref{RSC} and Jensen's inequality, along with some algebra, we have
\begin{align}
\label{GlobalRSC}
    & \mL\left(\btheta^*+\hat{\bDelta}\right)-\mL\left(\btheta^*\right)-\left\langle\nabla \mL\left(\btheta^*\right), \hat{\bDelta}\right\rangle \nonumber\\
& =\hat{\bDelta}^{\top} \nabla^2 \mL\left(\btheta^*+\gamma \hat{\bDelta}\right) \hat{\bDelta} \quad(\gamma \in(0,1)) \nonumber\\
& =\sum_{k=1}^K \frac{n_{S}}{N} \left(\hat{\bDelta}^{(k)}\right)^{\top}\hat{\bSig}^{(k)} \hat{\bDelta}^{(k)}+2 \sum_{k=1}^K \frac{n_{S}}{N} \left(\hat{\bDelta}^{(k)}\right)^{\top} \hat{\bSig}^{(k)} \hat{\bDelta}^{(0)}+\left(\hat{\bDelta}^{(0)}\right)^{\top}\left(\sum_{k=1}^K \frac{n_{S}}{N} \hat{\bSig}^{(k)} + \frac{n_{T}}{N}\hat{\bSig}^{(0)}\right) \hat{\bDelta}^{(0)} \nonumber \\
& =\sum_{k=1}^K \frac{n_{S}}{N}\left(\hat{\bDelta}^{(k)}+\hat{\bDelta}^{(0)}\right)^{\top} \hat{\bSig}^{(k)}\left(\hat{\bDelta}^{(k)}+\hat{\bDelta}^{(0)}\right)+\frac{n_{T}}{N} \left(\hat{\bDelta}^{(0)}\right)^{\top} \hat{\bSig}^{(0)} \hat{\bDelta}^{(0)} \nonumber \\
& \geq \sum_{k=1}^K \frac{n_{S} \alpha_k}{N}\left\|\hat{\bDelta}^{(k)}+\hat{\bDelta}^{(0)}\right\|_2^2+\frac{n_{T} \alpha_0}{N}\left\|\hat{\bDelta}^{(0)}\right\|_2^2 -\mR \left(\hat{\bDelta}\right) \nonumber\\
& \geq \alpha_{\min} \left\| \sum_{k=1}^K \frac{n_{S}}{N} \hat{\bDelta}^{(k)}+\hat{\bDelta}^{(0)} \right\|_2^2 - \mR \left(\hat{\bDelta}\right) \nonumber\\
&= \alpha_{\min} \left\| \wev\right\|_2^2 - \mR \left(\hat{\bDelta}\right)
\end{align}
with probability larger than $1 - c_{1}\exp (-c_{2}n_{T})$, where we define $\alpha_{\min} := \min_{0 \le k \le K} \alpha_k$ and 
$$
\mR \left(\hat{\bDelta}\right):=\sum_{k=1}^K \frac{n_{S} \beta_{k}}{N} \frac{\log p}{n_{S}}\left\|\hat{\bDelta}^{(k)}+\hat{\bDelta}^{(0)}\right\|_1^2+\frac{n_{T} \beta_0}{N} \frac{\log p}{n_{T}}\left\|\hat{\bDelta}^{(0)}\right\|_1^2.
$$

Notice that by triangle inequality, 
$$
\begin{aligned}
\mR \left(\hat{\bDelta}\right)
& = \sum_{k=1}^K \frac{\beta_k \log p}{N} \left\|\hat{\bDelta}^{(k)}+\hat{\bDelta}^{(0)}\right\|_1^2+\frac{ \beta_0 \log p}{N} \left\|\hat{\bDelta}^{(0)}\right\|_1^2 \\
& \leq \sum_{k=1}^K \frac{2 \beta_k \log p}{N} \left\|\hat{\bDelta}^{(k)}\right\|_1^2+\sum_{k=0}^K \frac{2 \beta_k \log p}{N}\left\|\hat{\bDelta}^{(0)}\right\|_1^2
\end{aligned}
$$

According to the restricted set of directions outlined in (\ref{WeightedRSD}), it holds that 
$$
\sum_{k=1}^K \lambda_k\left\|\hat{\bDelta}^{(k)}\right\|_1+\lambda_0\left\|\hat{\bDelta}^{(0)}\right\|_1 \leq
\sum_{k=1}^K \lambda_k\left\|\hat{\bDelta}^{(k)}\right\|_1+2\lambda_0\left\|\hat{\bDelta}^{(0)}\right\|_1 \leq 8 \lambda_0\left\|\wev_S\right\|_1+8 \sum_{k=1}^K \lambda_k h_{k}.
$$
with probability larger than $1 - c_{1}\exp (-c_{2}\log p)$.

So if we define $\beta_{\max} = \max_{0 \le k \le K} \beta_k$, by triangle inequality and the fact that $|S|=s$, we then have
\begin{align*}
\mR \left(\hat{\bDelta}\right) &\leq \frac{2 \beta_{\max}\log p}{N} \left( \sum_{k=1}^K \frac{\lambda^2_{k}}{\lambda^2_{k}} \left\|\hat{\bDelta}^{(k)}\right\|_1^2+(K+1) \frac{\lambda^2_{0}}{\lambda^2_{0}} \left\|\hat{\bDelta}^{(0)}\right\|_1^2 \right) \\
 &\leq  \frac{2 \beta_{\max}}{\lambda^2_{k}\wedge(\lambda^2_{0}/(K+1)) } \frac{\log p}{N} \left( \sum_{k=1}^K \lambda^2_{k} \left\|\hat{\bDelta}^{(k)}\right\|_1^2+ \lambda^2_{0} \left\|\hat{\bDelta}^{(0)}\right\|_1^2 \right)  \\ 
 &\leq \frac{2 \beta_{\max}}{\lambda^2_{k}\wedge(\lambda^2_{0}/(K+1)) } \frac{\log p}{N} \left( \sum_{k=1}^K\lk \left\|\hat{\bDelta}^{(k)}\right\|_1+ \lambda_{0} \left\|\hat{\bDelta}^{(0)}\right\|_1 \right)^2 \\
 & \leq \frac{2 \beta_{\max}}{\lambda^2_{k}\wedge(\lambda^2_{0}/(K+1)) } \frac{\log p}{N} 
 \left(8 \lambda_0\left\|\wev_S\right\|_1+8 \sum_{k=1}^K \lambda_k h_{k}\right)^2 \\
 & \leq \frac{2 \beta_{\max}}{\lambda^2_{k}\wedge(\lambda^2_{0}/(K+1)) } \frac{\log p}{N} 
 \left(128 \lambda^2_0\left\|\wev_S\right\|^2_1+128 \left(\sum_{k=1}^K \lambda_k h_{k}\right)^2 \right) \\
  & \leq \frac{2 \beta_{\max}}{\lambda^2_{k}\wedge(\lambda^2_{0}/(K+1)) } \frac{\log p}{N} 
 \left(128 s \lambda^2_0\left\|\wev\right\|^2_2+128 \left(\sum_{k=1}^K \lambda_k h_{k}\right)^2 \right)
\end{align*}
Recall that we introduce the shorthand $u_{n}=\frac{ 256\beta_{\max }\lz^2}{\alpha_{\min}\lambda^2_{k}\wedge(\lambda^2_{0}/(K+1)) }\frac{s \log p}{N} $, and $v_{n} = \frac{256 \beta_{\max}}{\lambda^2_{k}\wedge(\lambda^2_{0}/(K+1)) } \frac{\log p}{N} \left(\sum_{k=1}^{K}\lk h_{k}\right)$, combining the above argument with (\ref{GlobalRSC}) leads to
\begin{align}
\mL\left(\btheta^*+\hat{\bDelta}\right)-\mL\left(\btheta^*\right)-\left\langle\nabla \mL\left(\btheta^*\right), \hat{\bDelta}\right\rangle  \geq (1 - u_{n})\alpha_{\min} \left\| \wev\right\|_2^2-v_{n} \sum_{k=1}^K \lambda_k h_{k} 
\end{align}
with probability larger than $1 - c_{1}\exp (-c_{2}n_{T}) - c_{3}\exp (-c_{4}\log p)$, which finishes the proof.

\subsection{Proof of Lemma \ref{steponel1bound}}
The result in $\ell_{2}$-norm follows directly from the proof of Theorem \ref{stepone}. Here we investigate the upper bound for the estimation error in $\ell_{1}$-norm, i.e., $\|\wev\|_1$. We categorize the problem into two cases based on the relationship between $\frac{1}{4}\lambda_0 \left\|\wev\right\|_{1}$ and $2\sum_{k=1}^{K}\lk h_{k}$, and discuss by cases.

Starting with the first scenario where $\frac{1}{4}\lambda_0 \left\|\wev\right\|_{1} > 2\sum_{k=1}^{K}\lk h_{k}$, in such case (\ref{unifiedproperty2}) implies
$$
0 \geq \hat{\bDelta}^\top \hat{\bSig} \hat{\bDelta}-\frac{7}{4}\lambda_0 \left\|\wev\right\|_{1}+2\lambda_{0}\left\|\wev_{S^{c}}\right\|_{1} \geq -\frac{7}{4}\lambda_0 \left\|\wev\right\|_{1}+2\lambda_{0}\left\|\wev_{S^{c}}\right\|_{1},
$$
which implies $\frac{1}{4}\|\wev_{S^c}\|_1  \leq \frac{7}{4} \|\wev_{S}\|_1$, and $\|\wev\|_{1} \leq 8 \|\wev_{S}\|_{1} \leq 8\sqrt{s}\|\wev_{S}\|_{2} \leq 8\sqrt{s}\|\wev\|_{2}$. Therefore, in this case, we obtain
$$
\|\wev\|_{1} \lesssim s \lambda_{0} + \sqrt{s} \sqrt{\sum^{K}_{k=1} \lk h_{k}}
$$
with probability larger than $1 - c_{1}\exp(c_2 n_{T}) - c_{3} \exp(c_4 \log p)$.

We now transit to the second scenario when $\frac{1}{4}\lambda_0 \left\|\wev\right\|_{1} \le 2\sum_{k=1}^{K}\lk h_{k}$. In this instance, we directly obtain
$$
\|\wev\|_1 \leq \frac{8 \sum^{K}_{k=1} \lambda_{k}h_{k} }{ \lambda_{0}}
$$

Taking into account both the discussed scenarios, we can conclude that 
\begin{align}
\label{l1errorbound}
\|\wev\|_{1}& \lesssim s \lambda_{0} + \sqrt{s} \sqrt{\sum^{K}_{k=1} \lk h_{k}} + \frac{\sum^{K}_{k=1} \lk h_{k}}{\lz} 
\end{align}
with probability larger than $1 - c_{1}\exp(c_2 n_{T}) - c_{3} \exp(c_4 \log p)$, which is as desired.

\subsection{Proof of Lemma \ref{DebiasBound}}

We start with the first term, $\frac{1}{n_{S}} \hThetak\left(\bX^{(k)}\right)^{\top} \bek$. Let $\left(\boldsymbol{a}_j^{(k)}\right)^\top:=\boldsymbol{e}_j^\top \hThetak \left(\bX^{(k)}\right)^{\top}$. Similar to the proof of Lemma \ref{concentration}, by \cite{vershynin2010introduction}, Proposition 5.10 and a union bound, 
$$
P\left(\max_{1 \le j\le p}\left|\frac{1}{n_{S}} \left(\boldsymbol{a}_j^{(k)}\right)^\top \bek\right|>t  \ \middle| \ \left\{\boldsymbol{a}_{j}^{(k)}\right\}_{1 \le j \le p}\right) \leq p \exp \left(-\frac{c n_{S}^2 t^2}{\sigma_k^2 \max_{1 \le j \le p}\left\|\boldsymbol{a}_{j}^{(k)}\right\|_2^2}\right)
$$
for some universal constant $c>0$. Therefore, to prove (\ref{VarPart}), it suffices to bound
$$
c_{\Omega}:=\frac{1}{n_{S}}\max_{1 \le j \le p}\left\|\boldsymbol{a}_{j}^{(k)}\right\|_2^2=\max_{1 \le j \le p}\left(\hThetak \hat{\bSig}^{(k)} \left(\hThetak\right)^\top\right)_{j, j}.
$$
In order to accomplish this, we employ Lemma 23 from \cite{lee2017communication}, which is formulated as follows.

\begin{lemma}
\label{InverseConcentration}
Under the conditions of Lemma \ref{DebiasBound},
$$
P\left(\max _{1 \le j \le p} \left(\Sigk\right)^{-1}_{j}\hat{\bSig}^{(k)}\left(\Sigk\right)^{-1}_{j}>2 \max _{1 \le j \le p} \left(\Sigk\right)_{j, j}^{-1}\right) \leq 2 p e^{-c_1 n_{S}}
$$
for some universal constant $c_1>0$.
\end{lemma}

Since $\Theta^{(k)}$ is the solution of problem (\ref{JMalgorithm}), combining the optimally condition with Lemma \ref{InverseConcentration} implies
$$
\max_{1 \le j \le p}\left(\hThetak \hat{\bSig}^{(k)} \left(\hThetak\right)^\top\right)_{j, j} \le \max _{1 \le j \le p} \left(\Sigk\right)^{-1}_{j}\hat{\bSig}^{(k)}\left(\Sigk\right)^{-1}_{j} \le 2 \max _{1 \le j \le p} \left(\Sigk\right)_{j, j}^{-1}
$$
with probability at least $1-2 p e^{-c_1 n_{S}}$. By assumption \ref{A1}, $\left(\Sigk\right)_{j, j}^{-1}$ is bounded above. Therefore (\ref{VarPart}) holds.

Next, we aim at the second term, $\boldsymbol{b}^{(k)}$. Applying Hölder's inequality to each component we obtain

$$
\begin{aligned}
\left\|\boldsymbol{b}^{(k)}\right\|_{\infty} & =\left\|\left(\hThetak \hat{\bSig}^{(k)}-\boldsymbol{I}\right)\left(\hat{\bb}_{\LASSO}^{(k)}-\bk\right)\right\|_{\infty} \\
& \leqslant \max _{1 \leq j \leq p}\left\|\hat{\boldsymbol{\Theta}}_{j}^{(k)} \hat{\bSig}^{(k)}-\boldsymbol{e}_{j}^{\top}\right\|_{\infty}\left\|\hat{\bb}_{\LASSO}^{(k)}-\bk\right\|_{1}
\end{aligned}
$$

where recall that $\hat{\boldsymbol{\Theta}}_{j}^{(k)}$ denotes the $j$th row of $\hThetak$. By the optimality condition of (2) and the choice of $\mu_{k}$, we have
\begin{align*}
  \max _{1 \leq j \leq p}\left\|\hat{\boldsymbol{\Theta}}_{j}^{(k)} \hat{\bSig}^{(k)}-\boldsymbol{e}_{j}^{\top}\right\|_{\infty} \lesssim \sqrt{\frac{\log p}{n_{S}}}.
\end{align*}

In addition, recall that $\left\|\bb^{(0)}\right\|_{0}=s$ and $\left\|\bb^{(0)}-\bk\right\|_{1} \le h_{k}$, so we can show the following Lemma:
\begin{lemma}
    \label{weakbound}
    Under the condition of Lemma \ref{DebiasBound}, we have
$$
\left\|\hat{\bb}_{\LASSO}^{(k)}-\bk\right\|_{1} \leqslant s \sqrt{\frac{{\log p}}{n_{S}}}+h_{k}
$$
with probability at least $1-c_{1} p^{-c_{2}}$.
\end{lemma}

The proof of Lemma \ref{weakbound} comes from a direct application of Lemma 1 in \cite{li2022transfer}. Integrating the above arguments leads to $\left\|\boldsymbol{b}^{(k)}\right\|_{\infty} \lesssim_{P}  \frac{s\log p}{n_{S}}+h_{k}\sqrt{\frac{\log p}{n_{S}}}$ with probability larger than $1 - c_{1} \exp(c_2 \log p)$. This finishes the proof.

\subsection{Proof of Lemma \ref{ComConcentrastion}}
Following the proof of Lemma \ref{concentration}, by applying Hölder's inequality, we obtain
\begin{align*}
    \left|\left\langle\nabla \mL\left(\btheta\right), \bDelta\right\rangle\right| &=\sum_{k=1}^{K} \left| \left\langle\frac{\sqrt{n_{S}}}{N}  \tilde{\boldsymbol{\epsilon}}^{(k)}, \bDelta^{(k)}\right\rangle \right| +\left|\left\langle\sum_{k=1}^{K} \frac{\sqrt{n_{S}}}{N}  \tilde{\boldsymbol{\epsilon}}^{(k)} + \frac{1}{N} \left(\bXz\right)^{\top} \bez, \bDelta^{(0)}\right\rangle \right| \\
& \leq \sum_{k=1}^{K}\left\|\frac{\sqrt{n_{S}}}{N}  \tilde{\boldsymbol{\epsilon}}^{(k)}\right\|_{\infty}\left\|\bDelta^{(k)}\right\|_{1}+\left\|\sum_{k=1}^{K} \frac{\sqrt{n_{S}}}{N}  \tilde{\boldsymbol{\epsilon}}^{(k)} + \frac{1}{N} \left(\bXz\right)^{\top} \bez\right\|_{\infty}\left\|\bDelta^{(0)}\right\|_{1} \\
& \leq \sum_{k=1}^{K}\left\|\frac{1}{N} \hThetak\left(\bX^{(k)}\right)^{\top} \bek + \frac{n_{S}}{N}\boldsymbol{b}^{(k)}\right\|_{\infty}\left\|\bDelta^{(k)}\right\|_{1} \\
& \quad + \left\|\frac{1}{N}\left(\sum^{K}_{k=1} \hThetak\left(\bX^{(k)}\right)^{\top} \bek+ \left(\bXz\right)^{\top} \bez\right) + \sum^{K}_{k=1} \frac{n_{S}}{N}\boldsymbol{b}^{(k)}\right\|_{\infty}\left\|\bDelta^{(0)}\right\|_{1}
\end{align*}

We start with the first term on the right-hand side of the inequality. In Lemma \ref{DebiasBound} we have shown that with probability larger than $1- c_{1}\exp(-c_{2}\log p)$,
\begin{align}
\label{Reemphasize}
\frac{1}{n_{S}}\left\|\hThetak\left(\bX^{(k)}\right)^{\top} \bek\right\|_{\infty} \lesssim \sqrt{\frac{\log p}{n_{S}}} 
\text{\quad and \quad} 
\left\|\boldsymbol{b}^{(k)}\right\|_{\infty} \lesssim \frac{s \log p}{n_{S}}+h_{k} \sqrt{\frac{\log p}{n_{S}}}.
\end{align}
Thus in order to guarantee 
$$
\lambda_{k} \ge \left\|\frac{1}{N} \hThetak\left(\bX^{(k)}\right)^{\top} \bek + \frac{n_{S}}{N}\boldsymbol{b}^{(k)}\right\|_{\infty},
$$
it suffices to choose $\lambda_{k} = c_k \left( \sqrt{\frac{n_{S}}{N} \frac{\log p}{N}} + \frac{s \log p}{N} + \frac{n_{S}}{N} \sqrt{\frac{\log p}{n_{S}}}h_{k}\right)$ for some sufficiently large constant $c_{k}$. 

Next, we shift our focus to the second term. Similar to the arguments in Proposition \ref{DebiasBound}, we denote $\left(\boldsymbol{a}_j^{(k)}\right)^\top:=\boldsymbol{e}_j^\top\hThetak \left(\bX^{(k)}\right)^{\top}$ for $k=1,\dots,K$ and $\left(\boldsymbol{a}_{j}^{(0)}\right)^\top:=\boldsymbol{e}_{j}^\top \left(\bXz\right)^{\top}$. By \cite{vershynin2010introduction}, Proposition 5.10 and a union bound,
\begin{align*}
P\left(\max_{1 \le j\le p}\left|\frac{1}{N} \sum^{K}_{k=0} \left(\boldsymbol{a}_j^{(k)}\right)^\top \bek\right|>t \ \middle| \ \left\{\boldsymbol{a}_j^{(k)}\right\}_{1 \le j \le p, 0 \le k \le K}\right) \leq p \exp \left(-\frac{c N^2 t^2}{\max_{1 \le j \le p, 0 \le k \le K}\sigma_{k}^2\left\|\boldsymbol{a}_{j}^{(k)}\right\|_2^2}\right)
\end{align*}
According to Proposition \ref{DebiasBound} and Assumption \ref{A1}, $\max_{0 \le k \le K, 1 \le j \le p}\left\|\boldsymbol{a}_{j}^{(k)}\right\|_2^2$ is bounded above with probability larger than $1-c_{1} \exp(-c_{2}\log p)$. This result together with (\ref{Reemphasize}) indicates that 
$$
 \left\|\frac{1}{N}\left(\sum^{K}_{k=1} \hThetak\left(\bX^{(k)}\right)^{\top} \bek+ \left(\bXz\right)^{\top} \bez\right) + \sum^{K}_{k=1} \frac{n_{S}}{N}\boldsymbol{b}^{(k)}\right\|_{\infty} \lesssim \sqrt{\frac{\log p}{N}} + \frac{Ks \log p}{N} + \sum^{K}_{k=1} \frac{n_{S}}{N}\sqrt{\frac{\log p}{n_{S}}}h_{k}
$$
so it suffices to choose $\lambda_{0} = c_0 \left( \sqrt{\frac{\log p}{N}} + \frac{Ks \log p}{N} + \sum^{K}_{k=1} \frac{n_{S}}{N}\sqrt{\frac{\log p}{n_{S}}}h_{k} \right)$ for some constant $c_{0}$. This finishes the proof.

\section{Impact of Covariate Shift on $C_\Sigma$} \label{app:c_sigma}
Recall constant $C_\Sigma$ defined as
$$
C_{\Sigma}:=1+\max _{j \leq p} \max _{k}\left\|e_j^{\top}\left(\Sigk-\Sigz\right)\left(\sum_{1 \le k \le K}\frac{1}{K}\Sigk\right)^{-1}\right\|_1.
$$
Clearly, $C_\Sigma$ depends on the difference between the source and target covariance matrices. In the following, we provide an example where $C_{\Sigma}$ diverges with $p$.

Let  $K=1$, i.e., there is only one source dataset, and the target data covariance matrix $\Sigz = \mathbf{I}$. The source data covariance $\Sigk$ is constructed as $\Sigk = \alpha \mathbf{A} + (1 - \alpha) \mathbf{I}$, where $\mathbf{A}$ is an arrowhead matrix with all nonzero elements equal to one. Next, we provide the choice of $\alpha$ such that the eigenvalues of $\Sigk$ are bounded, thus satisfying Assumption~\ref{A1}. To this end, we first provide an expression of the eigenvalues of $\mathbf{A}$.
Applying Cauchy's interlace theorem, we have $\Lambda_{\min} (\mathbf{A} ) \leq 1$, $\Lambda_{\max} (\mathbf{A} ) \geq 1$, and all of the rest eigenvalues equal to one. Further using the fact that ${\rm Tr} (\mathbf{A} ) = p$ and ${\rm det} (\mathbf{A} ) = - (p-2)$ we conclude  $\Lambda_{\min} (\mathbf{A} ) = 1 - \sqrt{p-1}$ and $\Lambda_{\max} (\mathbf{A} ) = 1 + \sqrt{p-1}$. 
Based on the eigenvalues we set $\alpha = c/\sqrt{p-1}$ with constant $c \in (0,1)$, which gives $\Lambda_{\min} (\Sigk) = 1-c$ and $\Lambda_{\max} (\Sigk) = 1+c$. 

We then compute $C_\Sigma$ under the above setting. Using the formula for the inverse of arrowhead matrices provided  in~\cite{salkuyeh2018explicit}, we obtain
\begin{align}
   \left(\Sigk-\Sigz\right)\left(\sum_{1 \le k \le K}\frac{1}{K}\Sigk\right)^{-1} = \left(
\begin{array}{c|c}
  1-\beta^{-1} & \alpha\beta^{-1} \cdots \alpha\beta^{-1}\\ \hline
  \alpha\beta^{-1} & \raisebox{-15pt}{{\large\mbox{{$- \alpha^2\beta^{-1}$}}}} \\[-4ex]
  \vdots & \\[-0.5ex]
  \alpha\beta^{-1} &
\end{array}
\right)_{p \times p}, 
\end{align}
where $\beta: = 1 - (p-1) \alpha^2$. Substituting $\alpha = c/\sqrt{p-1}$ reveals $C_\Sigma = O(\sqrt{p})$.

\section{Implementation of TransFusion}\label{app:TransFusion-implementation}
In this section, we show that how the first step of TransFusion method can be  implemented using the Proximal Gradient Descent (PGD) algorithm via a change of variables. The second de-bias step (\ref{transferstep}) is a standard LASSO problem and there is a rich literature on efficient numerical solutions, see~\cite{li2018highly} and the references therein.

For the first co-traning step, notice that under the one-to-one variable transformation
$
\btheta:=(
(\bb^{(1)}-\bb^{(0)})^\top,
(\bb^{(2)}-\bb^{(0)})^\top,
\dots,
(\bb^{(K)}-\bb^{(0)})^\top,
\bb^{(0)})
$,
solving problem (\ref{obj}) is equivalent to solving 
\begin{equation}\label{p:w-LASSO}
\hat{\btheta} \in \underset{\btheta}{\operatorname{argmin}} \left\{\frac{1}{2N}\left\|\by-\bX\btheta\right\|_{2}^{2}+\lambda_{0}\sum^{K}_{k=0} a_{k} \left\| \btheta^{(k)} \right\|_{1}\right\},
\end{equation}
where $a_0 = 1$ and $\by$ and $\bX$ are defined in~\eqref{transrule}, recalled below for convenience: 
\begin{align*}
\by:=\left(\begin{array}{c}
\by^{(1)} \\
\by^{(2)} \\
\vdots \\
\by^{(K)} \\
\by^{(0)}
\end{array}\right) \  \bX:=\left(\begin{array}{ccccc}
\bX^{(1)} & 0 & \cdots & 0 & \bX^{(1)} \\
0 & \bX^{(2)} & \cdots & 0 & \bX^{(2)} \\
\vdots & \vdots & \vdots & \vdots & \vdots \\
0 & 0 & \cdots & \bX^{(K)} & \bX^{(K)} \\
0 & 0 & \cdots & 0 & \bX^{(0)}
\end{array}\right).
\end{align*} 
Problem~\eqref{p:w-LASSO} is a weighted LASSO problem \citep{zou2006adaptive}, and thus the proximal gradient descent algorithm can be directly applied. Given initialization $\btheta_0 \in \mathbb{R}^{(K+1)p}$ and proximal parameter $\gamma >0$, the PGD iteration reads:
\begin{align}\label{alg:PGD}
    \btheta_{t+1} = \argmin_{\btheta \in \mathbb{R}^{(K+1)p}} \left\langle \frac{1}{N}\bX^\top( \bX\btheta_t - \by ), \btheta - \btheta_t \right\rangle + \frac{1}{2 \gamma} \| \btheta - \btheta_t \|^2 + \lambda_{0}\sum^{K}_{k=0} a_{k} \left\| \btheta^{(k)} \right\|_{1}.
\end{align}
Notably, although calculating the gradient $\frac{1}{N}\bX^\top(\bX\btheta_t - \by)$ appears to involve multiplying a $(K+1)p \times (K+1)p$ matrix by a $(K+1)p$-dimensional vector,
using the sparse structure of $\bX$ it is easy to see it can be obtained via computing quantities $\bX^{(k)\top}\bX^{(k)} \bb_t^{(k)}$ and $\bX^{(k)\top} \by^{(k)}$ for $k = 0,\ldots,K$. Therefore per iteration it only involves the multiplication of a $p \times p$ matrix by a $p$-dimensional vector, and can be computed efficiently in parallel. The proximal mapping~\eqref{alg:PGD} can also be computed in closed form via soft-thresholding. 

\section{{Additional Simulation Results}}\label{app:suppl-sim}
\label{simulation}
\begin{figure}[htbp]
  \centering
  \includegraphics[width = 0.7\textwidth]{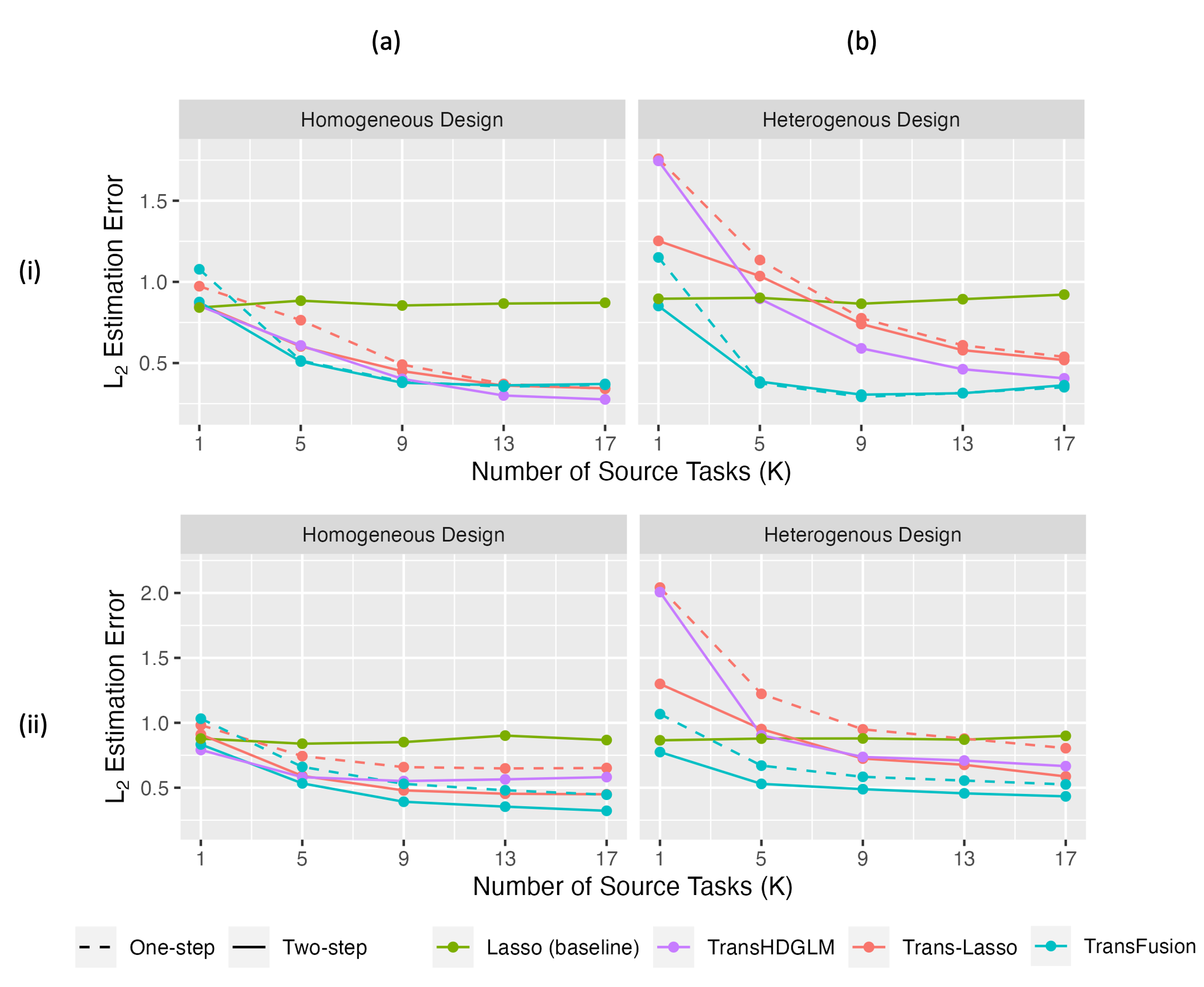}\vspace{-0.3cm}
\caption{Comparison of estimation errors under (i) diverse and (ii) non-diverse source task settings with (a) homogeneous design and (b) heterogeneous design with a large choice of $K$.}
  \label{fig:fourlargeK}
\end{figure}

\begin{figure}[htbp]
  \centering
  \includegraphics[width = 0.7\textwidth]{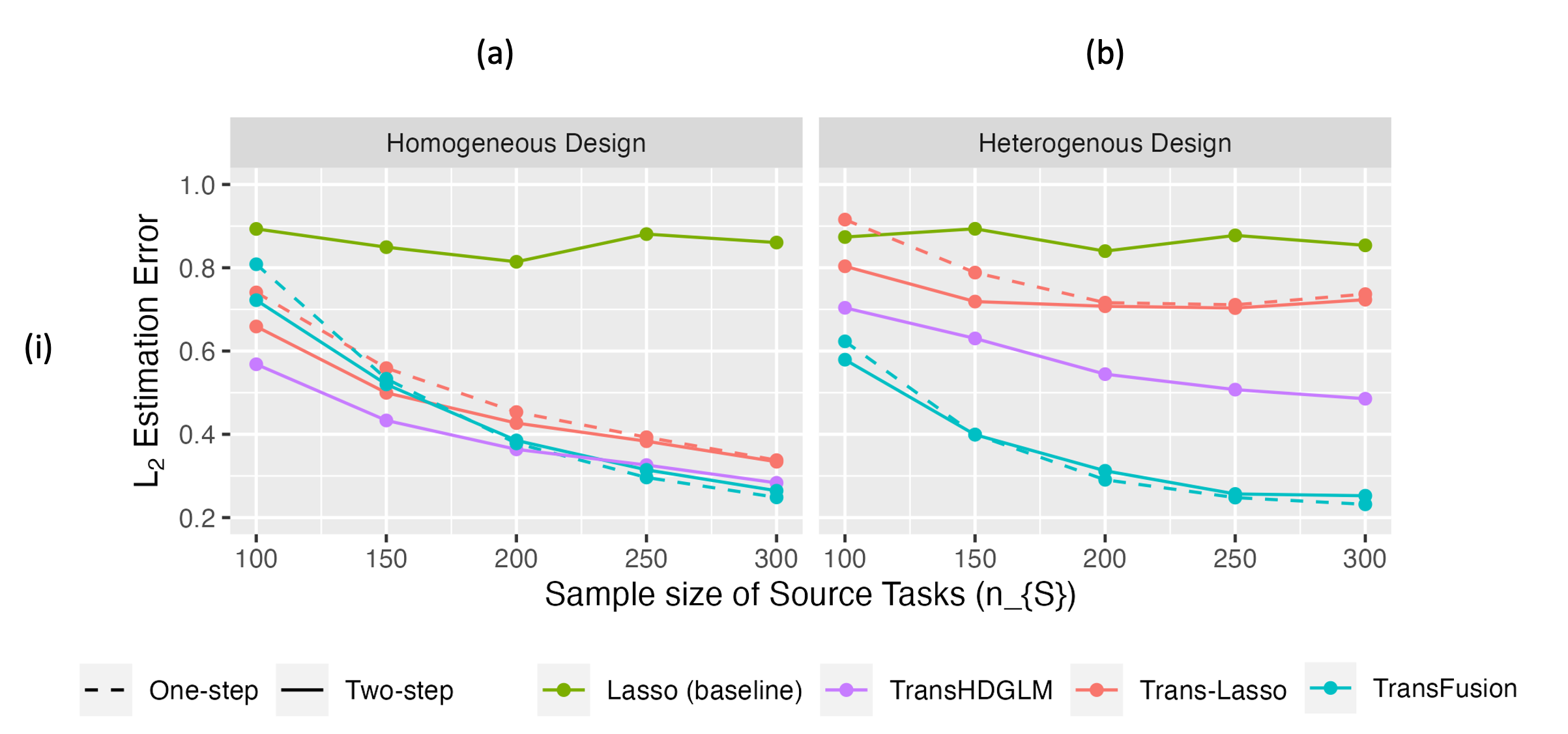}\vspace{-0.3cm}
\caption{Comparison of estimation errors under (i) diverse source task settings with (a) homogeneous design and (b) heterogeneous design with different choice of  $n_{S}$ and a fixed $K=10$.}
  \label{fig:fourchangeK}
\end{figure}

Fig. \ref{fig:fourlargeK} shows the results with the number of source tasks $K\in\{1,5,9,13,17\}$. The results are in general analogous to those in section \ref{sec:simulation}. One interesting observation is that in the diverse source task setting (i), as the source task number $K$ increases and the source task sample size $n_{S}$ remains a constant, Fig. \ref{fig:fourlargeK} (i-a) and (i-b) show diminishing improvement of TransFusion estimation error. This is because as our theoretical results suggest, in such settings, TransFusion achieves an estimation error of order $O(\frac{s \log p}{n_{T} + Kn_{S}} + (1+v_{n})\bar{h}\sqrt{\frac{\log p}{n_{T}}})$ with $v_{n}=\frac{K^2 \log p}{n_{S}}\bar{h}$. If we fix the source sample size $n_{S}$, increasing $K$ only decreases the first term, and the overall sum will be dominated by the second term. The term $v_n$ contributes to the U-shape error curve in the figures. This is the cost of using only local data to estimate the task-specific signal $\bdk$ in order to achieve robustness to covariate shifts.

Meanwhile, the theory implies that TransFusion would have a consistent error improvement if we proportionally increase the source sample size, $n_{S}$, with $K$. This is verified in Fig. \ref{fig:fourchangeK}, where, in the same diverse source task setting, a fixed $K$ with a growing $n_{S}$ results in a faster error reduction for TransFusion compared to other methods, especially in the existence of covariate shifts (Fig. \ref{fig:fourchangeK} (i-b)).

\section{Case Study: Handwritten-digit Classification}
\label{app:real-data}
We consider the problem of handwritten-digit classification based on the MNIST-C \citep{mu2019mnist} dataset. MNIST-C is a comprehensive suite of different corruptions applied to the MNIST dataset, for benchmarking out-of-distribution robustness in computer vision. This setup allows us to evaluate the covariate-shift robustness of the proposed \textit{TransFusion} algorithm. 

We choose images corrupted by ``brightness'', ``fog'' and ``motion blur'' as the source datasets ($K=3$), drawing $n_S$ source samples from each with $n_S \in \{500, 1000, 1500, 2000\}$. Then we set the original MNIST dataset as the target dataset, from which we collect $n_T = 100$ target samples. We use flattened pixel features of the images as features, amounting to $28 \times 28=784$ features per image ($p=784$). We transform the classification problem into 10 binary classification problems—one for each digit against all others, then evaluate the classification accuracy on $2000$ test images from the target dataset. 

For this multi-source binary classification task, we employ the \textit{TransFusion} algorithm and compare its performance with the \textit{TransLasso} algorithm and \textit{Lasso (baseline)} algorithm, all based on the logistic regression. Note that although in this paper we mainly focus on the linear regression setting, our theory and methodology can be easily generalized to the logistic regression setting~\citep{friedman2010regularization,negahban2012unified}. The implementation follows a similar manner as discussed in Appendix \ref{app:TransFusion-implementation}. 

\begin{figure}[htbp]
    \centering
    \includegraphics[width=1.0\linewidth]{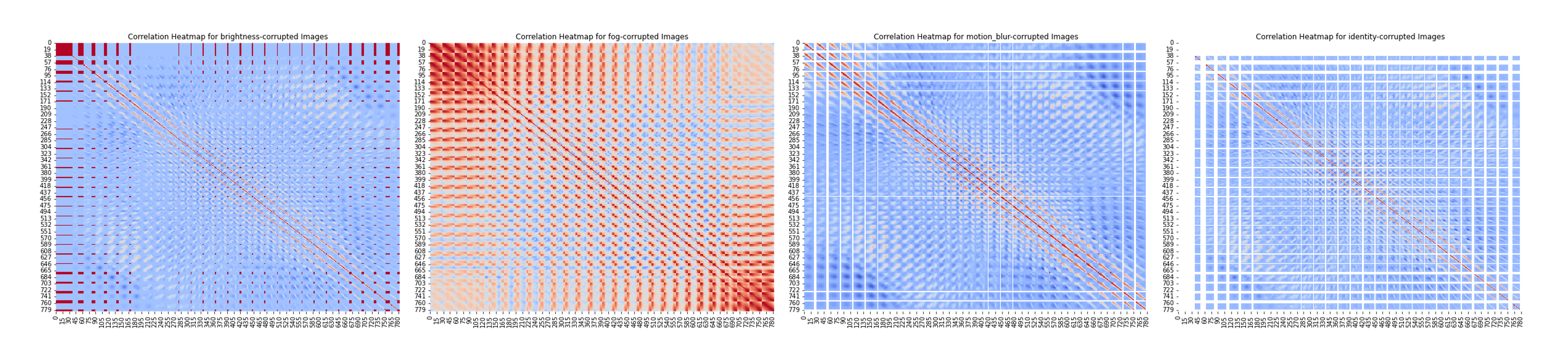}
    \caption{The correlation heatmaps of flattened pixel features for handwritten digit images affected by different types of corruptions. From left to right, the images are subjected to brightness corruption, fog corruption, motion blur corruption, and the original images without corruption (identity corruption).}
    \label{fig:cov_shift}
\end{figure}

One of the key challenges in this classification problem is managing the covariate shifts between different tasks. This was evident from the correlation heat maps of flattened pixel features in Figure \ref{fig:cov_shift}. Compared to the target sample, each source sample exhibits a distinct covariate correlation structure. Such covariate shifts should be carefully handled for effective knowledge transfer.

\begin{figure}[htbp]
    \centering
    \includegraphics[width=0.5\linewidth]{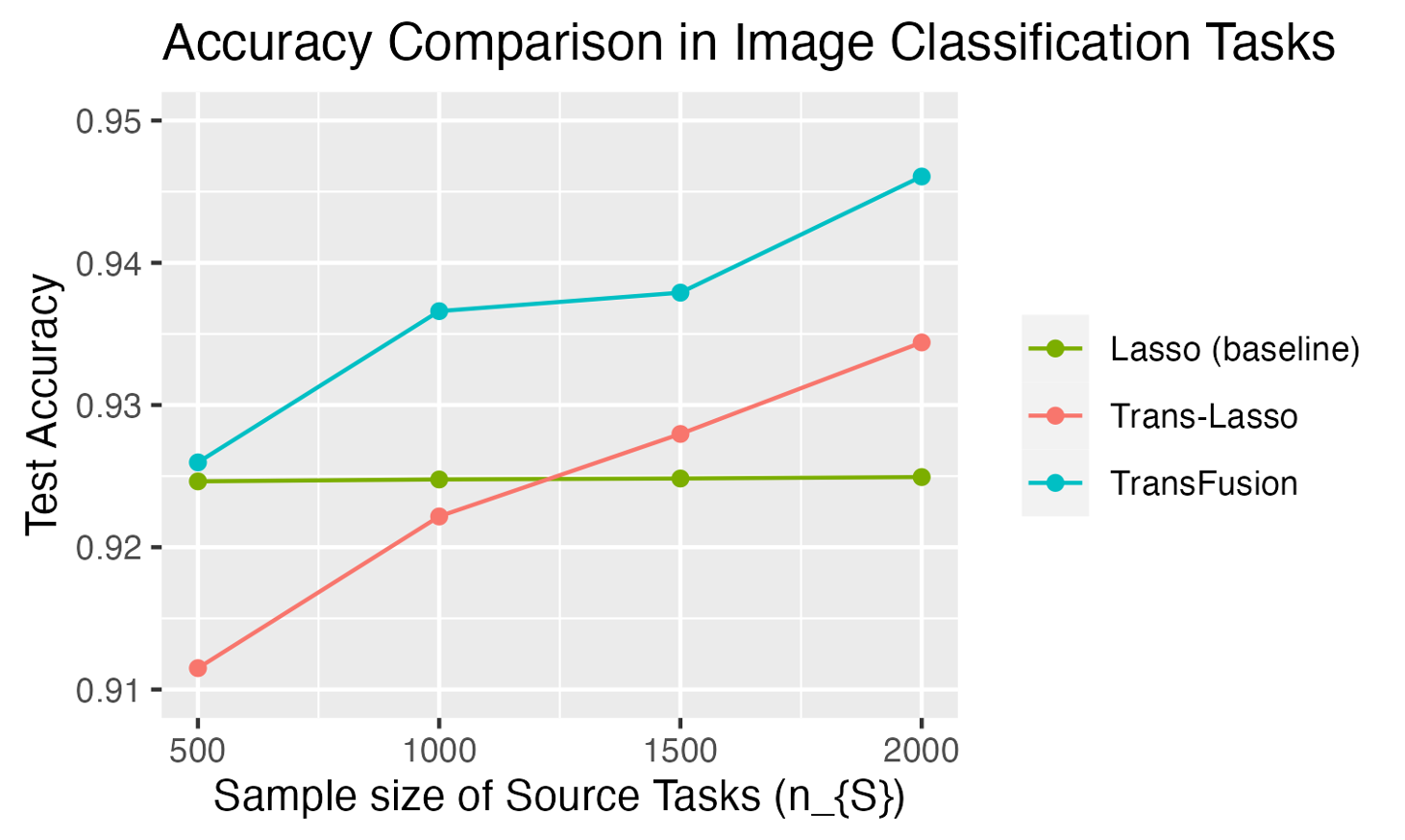}
    \caption{Test accuracy of different methods for the  binary digit classification problem averaged over 10 problems, plotted against varying source sample sizes $n_S$.}
    \label{fig:test_accu}
\end{figure}

Figure \ref{fig:test_accu} shows the average test accuracy across the 10 binary classification problems versus the source sample size $n_S$. From the figure, we can see that \textit{TransFusion} consistently outperforms other benchmarks. When the source sample size is relatively small, the \textit{Trans-Lasso} method even performs worse than the baseline Lasso method, suggesting that it fails to utilize the transferable knowledge under the covariate shift. In contrast, \textit{TransFusion} method still outperforms the baseline, indicating its robustness against the covariate shift. 

\section{Choice of $\tbk$}
\label{DebiasEst}
In this section, we specify how to choose $\tbk$.
 An intuitive option is the LASSO estimator computed based on source sample $(\bXk, \byk)$:
\begin{align}
\label{DebiasKKT}
\hbk_{\LASSO}\in\underset{\boldsymbol{\beta} \in \mathbb{R}^{(k+1)p}}{\operatorname{argmin}}\left\{\frac{1}{2n_{S}}\left\|\byk - \bXk \boldsymbol{\beta}\right\|_{2}^{2}+\tilde{\lambda}_{k}\|\boldsymbol{\beta}\|_{1}\right\}.
\end{align}
However, since the LASSO estimator is biased, computing  $\hat{\bw}_{C}$ by aggregating the local  $\hbk_{C}$s can only reduce the variance and has almost no effects on the bias \citep{mcdonald2009efficient}. To overcome such a drawback, we propose to first ``correct'' the bias  at the local level before transmitting it to the target node for transfer learning. This is achieved by debiasing $\hbk_{\LASSO} $ using the method proposed in~\citep{javanmard2014confidence}:

{\small
\begin{align}
\label{Comobj}
\tbk=\hbk_{\LASSO}+\frac{1}{n_{S}} \hThetak\left(\bX^{(k)}\right)^{\top}\left(\by^{(k)}-\bX^{(k)} \hat{\bb}^{(k)}_{\LASSO}\right).
\end{align}
}Here, $\hThetak$ serves as an approximation of $\left(\Sigk\right)^{-1}$, whose  $j$-th row is defined as the solution of the following optimization problem
\begin{align}
\label{JMalgorithm}
\begin{array}{cl}
\underset{\boldsymbol{\theta}_{j} \in \mathbb{R}^p}{\operatorname{minimize}} & \boldsymbol{\theta}_{j}^\top \hat{\boldsymbol{\Sigma}}^{(k)} \boldsymbol{\theta}_{j} \\
\text { subject to } & \left\|\hat{\boldsymbol{\Sigma}}^{(k)} \boldsymbol{\theta}_{j}-\boldsymbol{e}_j\right\|_{\infty} \leq \mu_{k} ,
\end{array}
\end{align}
with parameter $\mu_k >0$ properly chosen. The source code for solving the problem can be found at \url{https://web.stanford.edu/~montanar/sslasso/code.html}.

To understand the choice of $\tilde{\bb}$, we may rewrite~\eqref{Comobj} by subtracting $\bk$ from both sides to obtain
\begin{align}
\label{biasvarcomp}
\tbk-\bk=\frac{1}{n_{S}} \hThetak\left(\bX^{(k)}\right)^{\top} \boldsymbol{\epsilon}^{(k)} \quad -\left(\hThetak \hat{\boldsymbol{\Sigma}}^{(k)}-\boldsymbol{I}\right)\left(\hat{\bb}^{(k)}_{\LASSO}-\bk\right).
\end{align}
The first term on the right-hand side of the equation is associated with the variance of $\tbk$. Through (\ref{JMalgorithm}), we effectively minimize this variance. The second term, $\boldsymbol{b}^{(k)}$, on the other hand, contributes to the bias. By selecting an appropriate constraint parameter $\mu_{k}$ on $\|\hat{\boldsymbol{\Sigma}}^{(k)} \boldsymbol{\theta}_{j}-\boldsymbol{e}_j\|_{\infty}$ in (\ref{JMalgorithm}), we can control the bias term to be comparable or even smaller than the variance term, thereby mitigate the impact of bias on the later aggregation step. Hence, this choice of $\tilde{\bb}$ guarantees that the D-TransFusion could achieve much less communication overhead, while at the same time achieving the minimum performance loss compared to centralized TransFusion.



\end{document}